\documentclass[sigplan,10pt]{acmart}

\acmISBN{} 
\acmDOI{} 
\startPage{1}

\setcopyright{none}

\usepackage{booktabs}   
\usepackage{subcaption} 

\usepackage{algorithm}
\usepackage{xcolor}
\usepackage{amsmath}
\usepackage[noend]{algpseudocode}
\usepackage{algorithmicx}
\usepackage{appendix}
\usepackage{multirow}
\usepackage{diagbox}
\begin{document}

\title[Short Title]{FlashGS: Efficient 3D Gaussian Splatting for Large-scale and High-resolution Rendering} 


\author{
    Guofeng Feng$^{\dag}$, \,
    Siyan Chen$^{\dag}$, \,
    Rong Fu$^{*}$, \,
    Zimu Liao, \,
    Yi Wang, \,
    Tao Liu, \,
    Zhilin Pei, \,
    Hengjie Li, \,
    Xingcheng Zhang, \,
    Bo Dai
    \vspace{5pt}
\\
{\it Shanghai AI Laboratory}
}




\begin{abstract}
This work introduces FlashGS, an open-source CUDA Python library \footnote{Available at \href{https://github.com/InternLandMark/FlashGS}{https://github.com/InternLandMark/FlashGS}.} 
designed to facilitate the efficient differentiable rasterization of 3D Gaussian Splatting through algorithmic and kernel-level optimizations. FlashGS is developed based on the observations from a comprehensive analysis of rendering process to enhance computational efficiency and bring the technique to wide adoption. 
The paper includes a suite of optimization strategies, encompassing redundancy elimination, efficient pipelining, refined control and scheduling mechanisms, and memory access optimizations, all of which are meticulously integrated to amplify the performance of the rasterization process.
An extensive evaluation of FlashGS' performance has been conducted across a diverse spectrum of synthetic and real-world large-scale scenes, encompassing a variety of image resolutions. The empirical findings demonstrate that FlashGS consistently achieves an average 4x acceleration over mobile consumer GPUs, coupled with reduced memory consumption. 
These results underscore the superior performance and resource optimization capabilities of FlashGS, positioning it as a formidable tool in the domain of 3D rendering.
\end{abstract}



\maketitle 

\begingroup\renewcommand\thefootnote{*}
\footnotetext{Corresponding Author (e-mail: furong@pjlab.org.cn).}
\endgroup

\begingroup\renewcommand\thefootnote{\dag}
\footnotetext{The first two authors contribute equally to this work, and this work was done when they were interns at Shanghai AI Lab.}
\endgroup
\pagestyle{plain} 

\section{Introduction}
Neural Radiance Fields (NeRF)~\cite{Mildenhall2020NeRFRS}  have become popular for producing realistic image renderings, but the process of sampling numerous points per ray for pixel rendering hinders real-time capabilities. Recently, a new representation method known as 3D Gaussian Splatting (3DGS)~\cite{3DGS} has come to the forefront as a viable alternative. This technique has demonstrated the ability to attain real-time rendering speeds. The emergence of 3DGS has opened up a realm of possibilities, allowing for real-time exploration of indoor environments. This has far-reaching implications across various practical applications, including immersive free-viewpoint navigation, virtual property tours, and interactive virtual-reality gaming experiences.

\begin{figure}[htbp]
    \centering
    \includegraphics[width=1.0\linewidth]{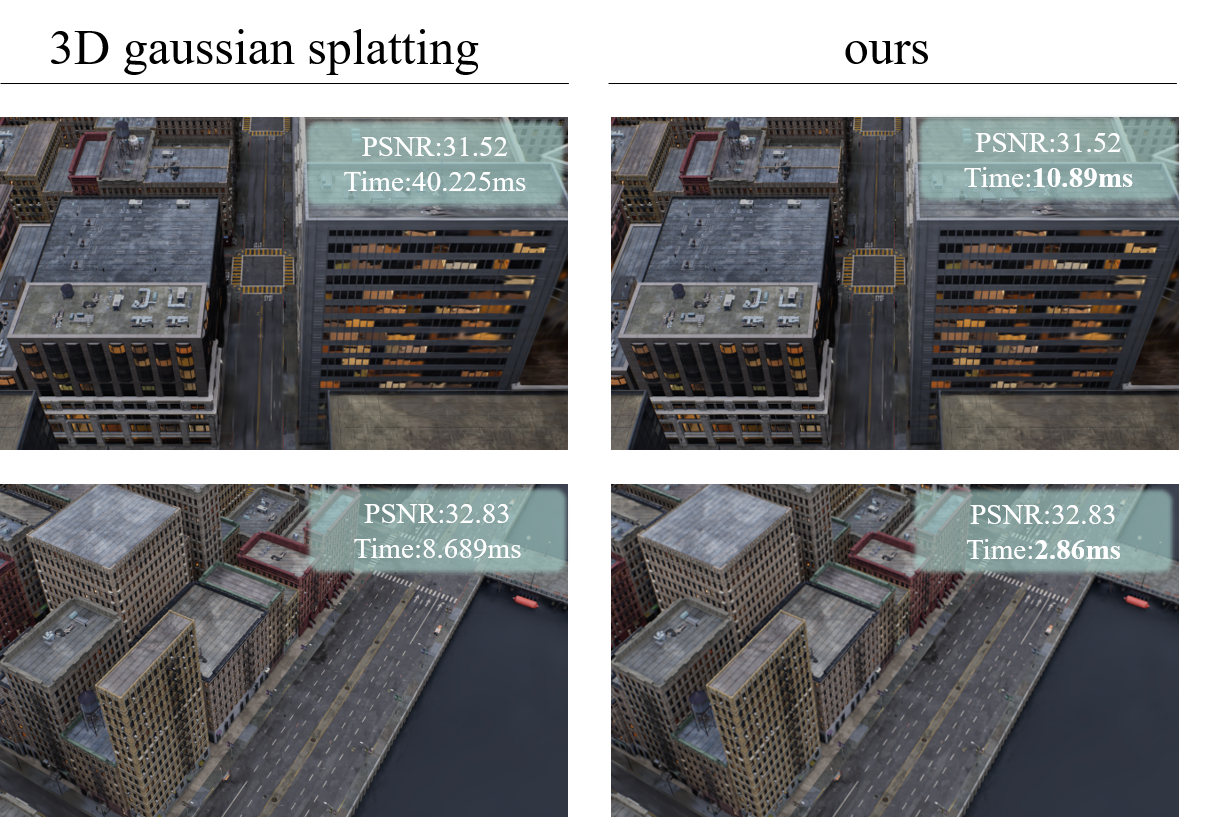}
    \caption{Two representative rendering output images with 3D Gaussian Splatting~\cite{3DGS} and our FlashGS. }
    \label{fig:result}
\end{figure}

Despite 3DGS's advantages, real-time rendering of large-scale or high-resolution areas on city-scale scenes~\cite{li2023matrixcity} or high quality scenes recorded by consumer GPS receivers~\cite{crandall2011discrete} is still hindered by limited computational and memory resources. 
This is due to the increase in the number of Gaussians and the size of each Gaussian as the image scale and resolution increase.

Existing works widely employ well-established compression or pruning methods to avoid storing or computing excessive Gaussians~\cite{girish2023eagles, lu2024scaffold, fan2023lightgaussian}. 
However, these methods have not yielded significant performance improvements. 
A recent work, GScore~\cite{lee2024gscore}, attempts to analyze the original 3DGS algorithm but primarily addresses limitations on mobile GPUs by designing a novel domain-specific hardware.

To address the shortcomings of current works, we've analyzed the 3DGS rendering process on a single consumer-grade GPU, identifying rendering performance bottlenecks, such as the rough intersection testing between Gaussians and tiles during preprocessing, and the massive redundant computations in volume rendering.
We further conduct a detailed performance profiling and extract several key factors hindering performance:
 1) Some Gaussian-tile pairs generated during preprocessing stage are not used in the actual rendering process.
 2) There are severe compute and memory access bottlenecks at different stages of the entire rasterization algorithm.

Based on our in-depth observations, we propose FlashGS, a novel 3DGS rendering algorithm with efficient system-level optimizations. Specifically, we first introduce a new design of rasterization workflow, constructing a more efficient rendering pipeline. 
For the implementation of this new rendering process, we firstly perform geometric and algebraic simplifications to alleviate high computational costs. We propose new runtime scheduling strategies to utilize the hardware, including the load balance and divergence elimination between threads, and a two-step prefetching execution pipeline to overlap computation and memory access operations. 
We also optimize memory management to better leverage the GPU memory hierarchy. 
Additionally, we further improve the performance by directly applying assembly-level optimizations.
Compared to 3DGS~\cite{3DGS}, GScore~\cite{lee2024gscore} and gsplat~\cite{ye2023mathematical}, FlashGS achieves up to 4$\times$ speedup and 49\% memory reduction, while keeping the same high quality metric peak signal-to-noise ratio (PSNR).

To the best of our knowledge, this is the first work to in-depth analyze, delicately re-design, and efficiently implement 3DGS-based volume rendering on a single consumer GPU. 
We make it possible to achieve efficient rendering for large-scale and high-resolution scenes with low overhead. 
Our main contributions are as follows:
\begin{itemize}
    \item We conduct an in-depth study of the original 3DGS algorithm and the current state-of-the-art research. We identify main challenges in large-scale and high-resolution scenes and pinpoint the fundamental problems of performance bottlenecks.
    \item We propose a new algorithm, FlashGS, which includes a precise redundancy elimination algorithm and an efficient rendering execution workflow.
    \item We systematically implement FlashGS on GPU with holistic optimizations, including computation, memory management, and scheduling.
    \item We test FlashGS on representative datesets, demonstrating that it significantly improves rendering speed while maintaining high image quality and low memory usage.
\end{itemize}

\section{Background and Related Work}
We briefly overview the novel-view synthesis and some existing methods, particularly the radiance field methods. Then, we introduce the current state-of-the-art, 3D Gaussian Splatting (3DGS). Finally, we briefly report some recent improvements in 3DGS.
\subsection{Novel View Synthesis}
\textbf{\textit{Novel view synthesis}} is an important method in 3D reconstruction. It generates new images from new viewpoints (target poses) based on a set of input images of a 3D scene from certain viewpoints (source poses). 
Before reconstruction, some representations are proposed to construct the 3D scene. Traditional methods use point clouds, voxels, or meshes to describe the spatial structure. 
However, these traditional explicit and discrete representations can lead to information loss and thus result in an incomplete representation of the scene.

\textbf{\textit{Neural Radiance Fields (NeRF)}}~\cite{Mildenhall2020NeRFRS} is a widely adopted attractive representation method, leveraging deep learning techniques. It implicitly models 3D scenes using multi-layer perceptrons (MLPs), and this continuous function achieves a fine representation of the scene. Some subsequent works focus on improving the quality of synthesized images through optimizing the models. Mip-NeRF~\cite{barron2021mip} is a multiscale model that casts cones and encodes the positions and sizes of frustum to address the aliasing issues of NeRF. Some methods~\cite{deng2022depth,wei2021nerfingmvs,xu2022point} leveraging depth supervision of point clouds to make the model converge faster. 
The substantial computation required by the MLP in NeRF results in slow training and rendering processes, which has inspired plenty of works on efficient training and inference. A common technique is storing the precompute training results to simplified data structures, such as sparse voxel grid~\cite{hedman2021baking} and octrees~\cite{yu2021plenoctrees,hu2022efficientnerf,wang2022fourier}. Other works propose or integrate new features with MLPs to enable faster training or inference, such as features in a voxel grid~\cite{liu2020neural,wu2022diver}. A notable work greatly accelerates NeRF, Instant-NGP~\cite{muller2022instant}, introduces a multi-resolution hash encoding grid, and simultaneously trains it with optimized MLPs.

\begin{figure*}[t]
    \centering
\includegraphics[width=\linewidth]{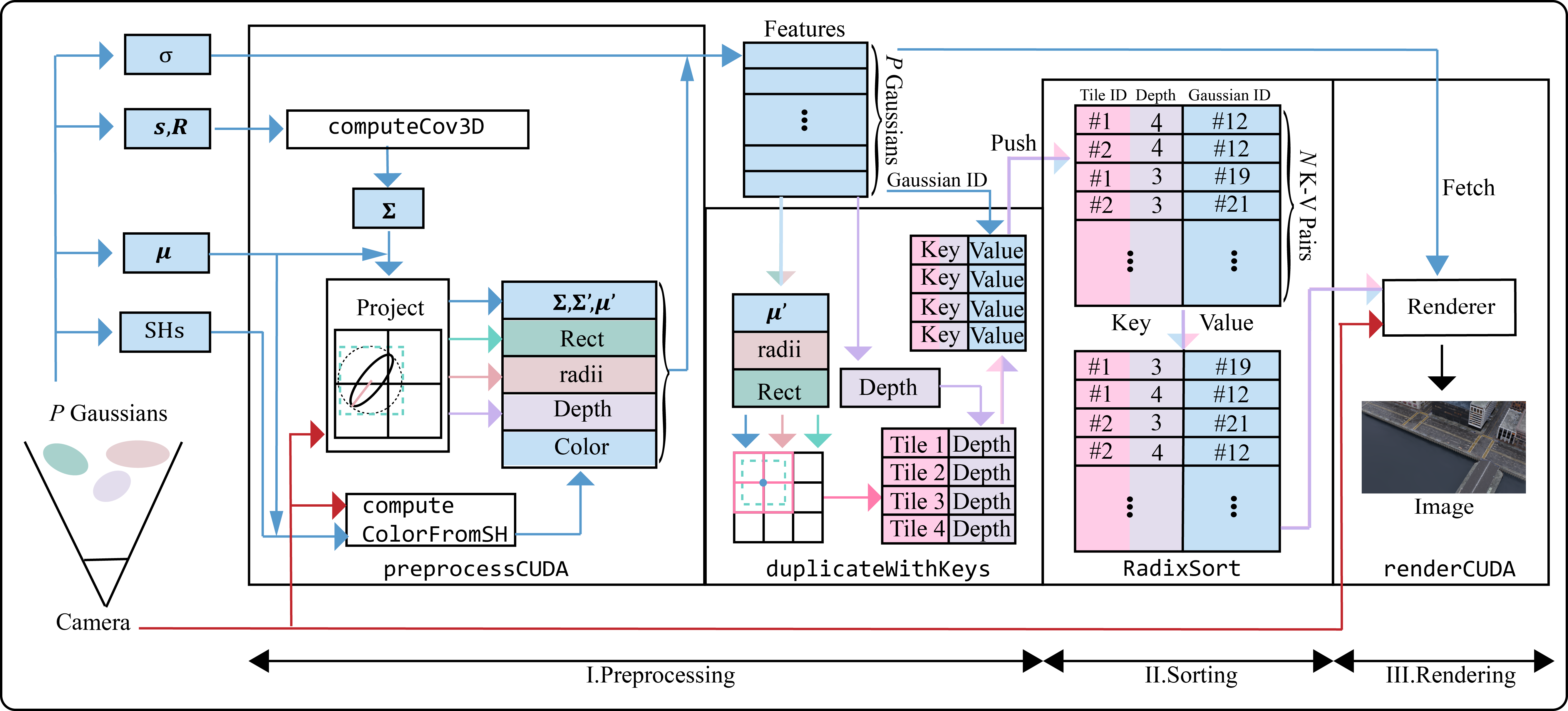}
    \caption{3DGS Overview}
    \label{fig:3DGS Overview}
\end{figure*}

\subsection{3D Gaussian Splatting}
\textbf{\textit{3D Gaussian Splatting (3DGS)}}~\cite{3DGS} models the scene with a set of Gaussian ellipsoids, allowing fast rendering through rasterizing these Gaussians into high quality images. Figure~\ref{fig:3DGS Overview} illustrate the 3DGS mechanism. The rasterizer can be coarsely divided into 3 steps: preprocessing (preprocessCUDA and DuplicateWithKeys), sorting and rendering. 

The preprocessing step in the original implementation contains two main operations. 
To begin with, $P$ 3D Gaussians after frustum culling are projected into the image plane as 2D Gaussians with mean $\boldsymbol\mu' \in \mathbb{R}^2$ and anisotropic covariance $\boldsymbol\Sigma' \in \mathbb{R}^{2\times2}$. 
Then the preprocessing pipeline assigns an ellipse with its axis-aligned bounding box (AABB) to each corresponding 2D Gaussian. 
3DGS uses AABB to get the intersected tiles with the ellipse, and for an AABB assigned by a certain Gaussian, the duplicateWithKeys operation traverses all tiles overlapping the AABB and organizes every tile with the Gaussian as key-value pairs. The keys are composed by the tile indexes and the depths of Gaussians, the values are the Gaussian indexes.
The sorting process sorts $N$ key-value pairs by depth for the front-to-back rendering, using the RadixSort in Nvidia CUB library. 
The final rendering precess obtains the pixel color taking the time complexity of $O(N)$ (the workload is divided into tiles) and each pixel color is computed considering the opacity $\alpha$ and transmittance.

\subsection{Improvement of 3DGS}
Although 3DGS has achieved a significant leap in 3D reconstruction, there remains considerable room for improvement. 
Some work has further improved image quality (photorealistic) by enhancing the sampling process of 3DGS, to eliminate blur and artifacts. 
Yan \textit{et al.}~\cite{yan2024multi} analyze the aliasing effect in 3DGS rendering and introduce a multi-scale 3D Gaussians representation to render different levels of detail of the scene, by improving the sampling frequency. Zhang \textit{et al.}~\cite{zhang2024fregs} propose FreGS, a novel framework leveraging regularization to perform coarse-to-fine high-quality Gaussian densification, to alleviate the over-construction problem. These aspects are not the focus of this paper, readers can find more details in some surveys~\cite{chen2024survey,fei20243d,wu2024recent}. 

In some high-resolution, large-scale scenes, 3DGS can generate millions of or even more (huge) Gaussians, putting immense pressure on memory units. This has inspired some works focusing on optimizing memory usage, which can be categorized into two main types: 1) Leveraging widely used techniques in model compression, like pruning and encoding. Scaffold-GS~\cite{lu2024scaffold} leverages the underlying scene geometry which uses anchor points growing and pruning strategies to effectively reduce redundant Gaussians, without loss of quality when rendering.
EAGLES~\cite{girish2023eagles} applies vector quantization (VQ) to quantize per-point attributes to compress the 3D Gaussian point clouds.
LightGaussian~\cite{fan2023lightgaussian} further introduces data distillation to reduce the degree of spherical harmonics coefficients and proposes a VQ strategy by the global significance of Gaussians.  
2) Introducing multi-GPU parallelism to avoid the memory constraints of a single GPU. 
Grendel~\cite{zhao2024scaling} is a scalable (up to 32 GPUs) distributed 3DGS training system, splitting 3DGS parameters across multiple GPUs and using sparse communication and dynamic load balance mechanisms. 
Chen \textit{et al.}~\cite{chen2024dogaussian} proposed DoGaussian, which decomposes a scene into K blocks and applies Alternating Direction Methods of Multipliers (ADMM) to train 3DGS distributedly.

However, only a handful of papers focus on the design and implementation of the original 3DGS algorithm itself and try to efficiently support critical operators in the computation process. 
This forms the direct motivation for our work. \textit{gsplat}~\cite{ye2023mathematical} is a CUDA-accelerated differentiable rasterization library for 3DGS, achieving faster and more memory efficient rendering. Durvasula \textit{et al.}~\cite{durvasula2023distwar} perform a performance characterization on the rasterization-based differential rendering, including 3DGS, and introduce a novel primitive using warp-level reduction to accelerate costly atomic operations during the gradient computation.
GScore~\cite{lee2024gscore} is a specific hardware acceleration unit to efficiently support the rendering pipeline over the mobile GPU along with some algorithmic co-design, based on an analysis of Gaussian-based rendering.

\section{Observation and Motivation}
\begin{figure}[htbp]
    \centering
    \includegraphics[width=\linewidth]{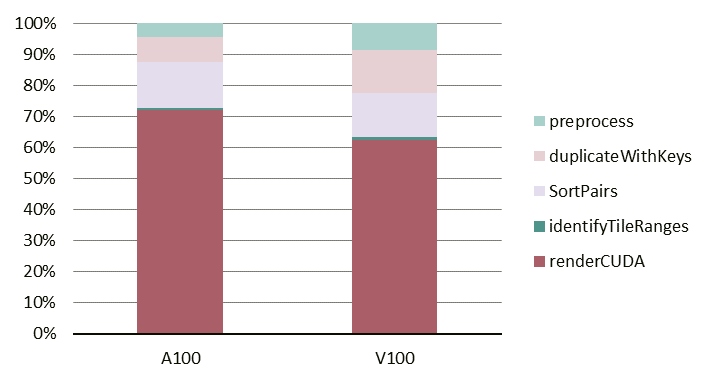}
    \caption{Runtime breakdown of 3GDS rasterization on the MatrixCity\cite{li2023matrixcity} dataset.}
    \label{fig:profile_chart}
\end{figure}

We conduct a detailed analysis of the 3D Gaussian Splatting rendering process. First, we obtain the critical steps in the rendering process by profiling the runtime breakdown. 
Based on it, we identify several significant performance issues, which inspired our optimization design.
As shown in the Figure~\ref{fig:profile_chart}, we test 3D Gaussian Splatting on A100 and V100 GPUs with a representative large-scale dataset, MatrixCity~\cite{ye2023mathematical}. 
From top to bottom it shows the time proportion of key operators in the rendering process. 
The rendering (renderCUDA) is the primary performance bottleneck, accounting for about 60\% of the total time. Gaussian sorting (SortPairs) and the preceding generation of the unsorted key-value list (DuplicateWithKeys) take up nearly 20\% of the time. Preprocessing time is within 10\%, but it also should not be overlooked. Given that these steps are not independent, we conduct a comprehensive study of them.

\subsection{Observation 1: Excessive redundant computations hinder efficient rendering.\label{sec:ob1}}
\begin{figure}[htbp]
    \centering
    \includegraphics[width=\linewidth]{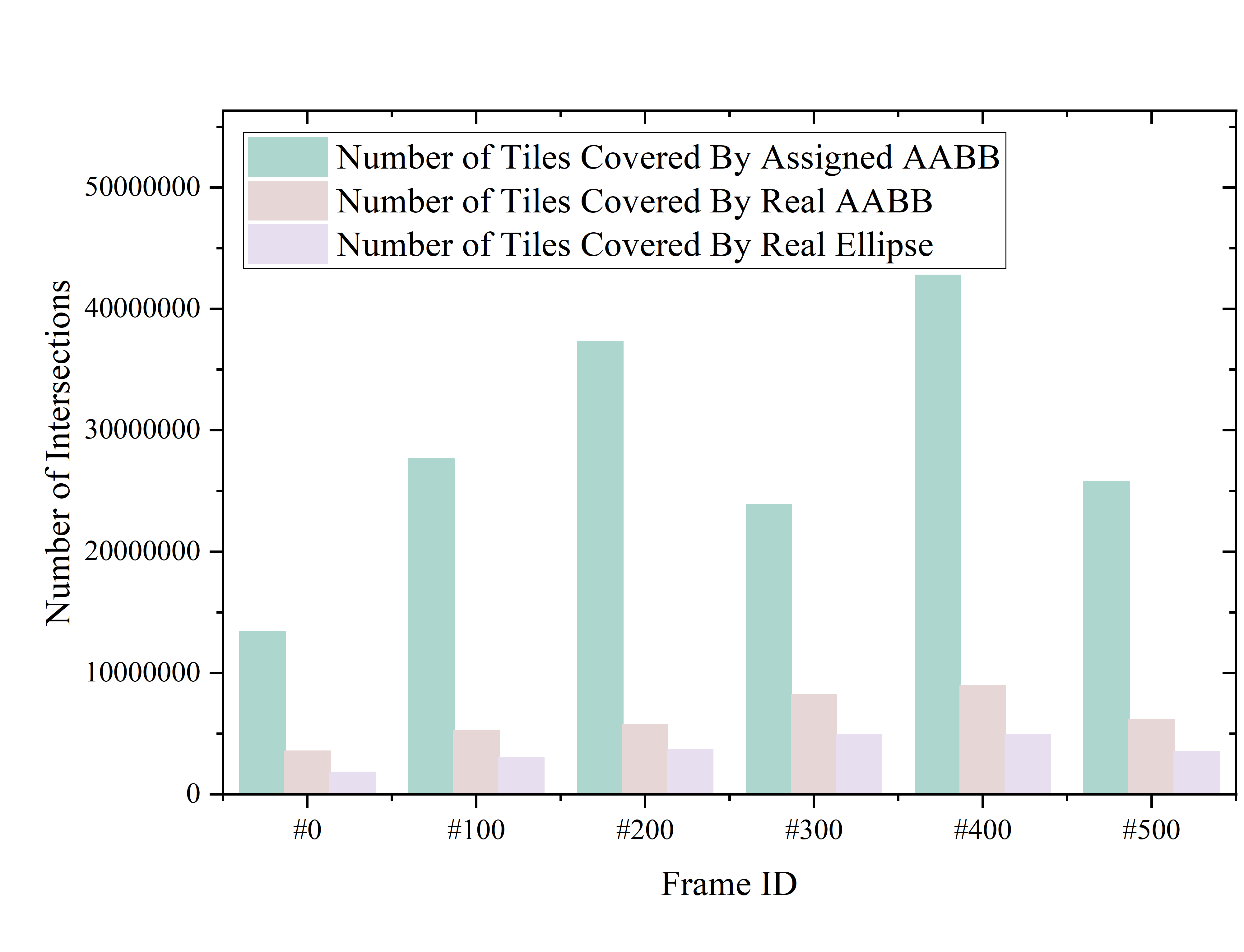}
    \caption{We evaluate the key-value pairs binning process from the rendering process of 6 frames in the scene trained from MatrixCity\cite{li2023matrixcity} dataset. The number of assigned k-v pairs is much more than the number of tiles really covered by the AABB or the projected ellipse.}
    \label{fig:redundancy_barchart}
\end{figure}

\begin{figure}[htbp] 
    \centering 
    \includegraphics[width=\linewidth]{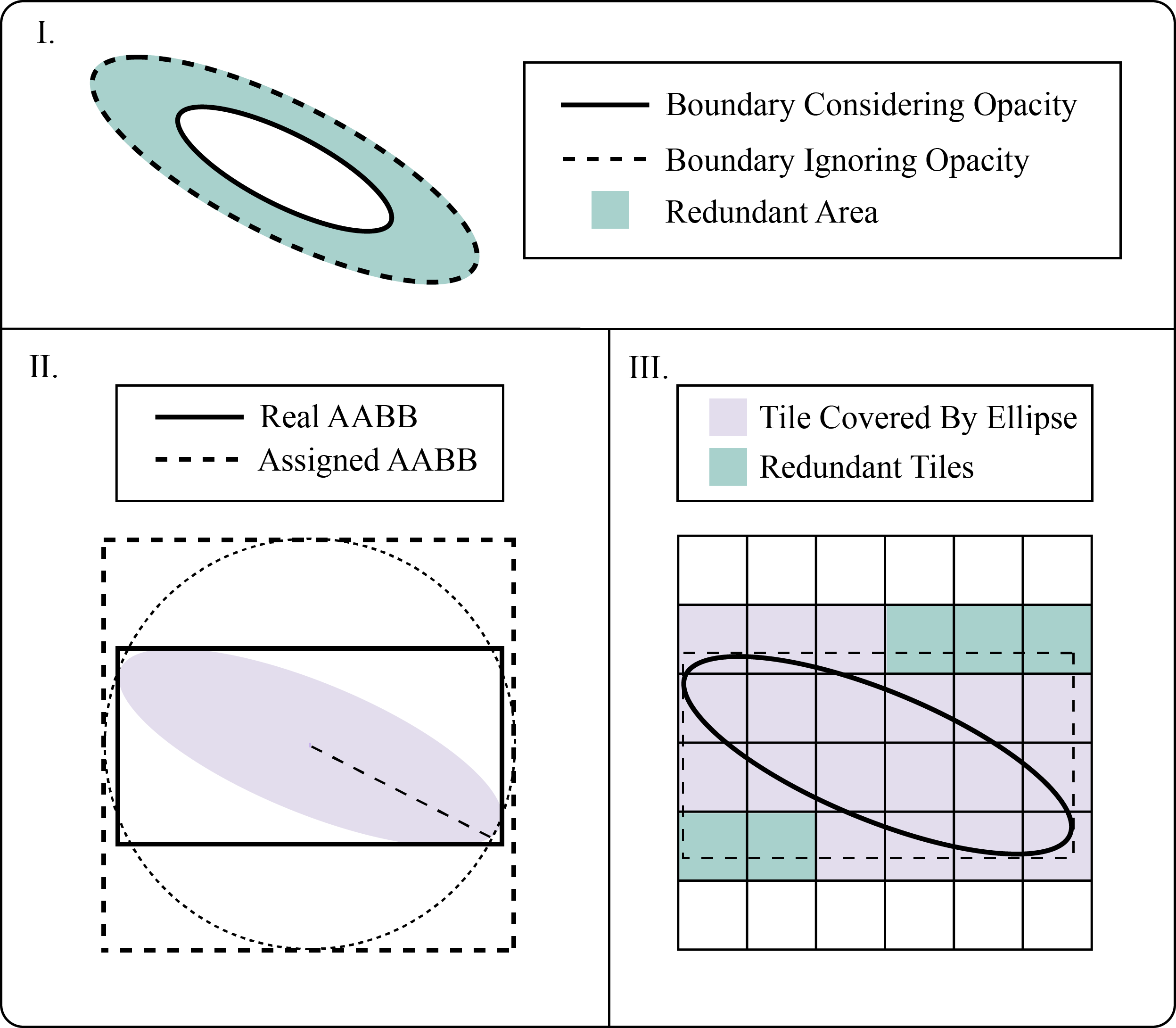} 
    \caption{Geometry Redundancies. There are 3 kinds of redundancies in original 3DGS intersection algorithm: I. The definition of ellipse ignores the opacity. II. The AABB is over-estimated. III. The tiles out of the ellipse are binned with the Gaussian.} 
    \label{fig:redundancy_overview} 
\end{figure}

We observe that there are several computational redundancies in the rasterizing pipeline. Figure \ref{fig:redundancy_barchart} shows that only a small proportion of the key-value pairs contribute to the result of rendering. 
Namely, the preprocessing kernels assigned unnecessary tiles without overlapping the projected ellipse to the Gaussian, creating redundant key values and inefficiencies in the sorting and rendering process. 
There are three main factors contribute to the redundancies as Figure \ref{fig:redundancy_overview} shows. 
We find that the redundancy I \& II in Figure \ref{fig:redundancy_overview} results in that the assigned AABB covers more tiles of the real AABB as Figure~\ref{fig:redundancy_barchart} shows. The redundancy III results in the difference between the number of tiles covered by the AABB and the projected ellipse.

\begin{figure}[htbp] 
    \centering 
    \includegraphics[width=.8\linewidth]{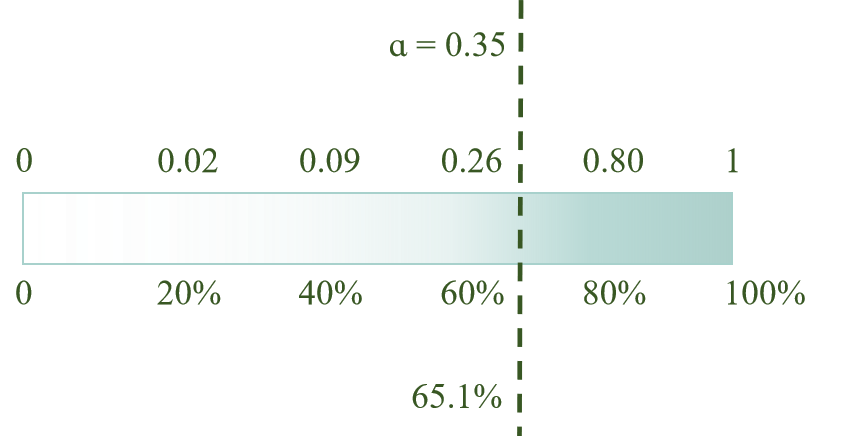} 
    \caption{Opacity distribution in the smallcity scene of MatrixCity dataset. The horizontal axis below the bar shows percentage ranging from 0\% to 100\%. These percentages correspond to the opacity values from 0 to 1 shown above the bar.}
    \label{fig:Opacity_distribution} 
\end{figure}
\paragraph{The opacity of Gaussians is ignored when defining the ellipse.} The original implementation adopts the three-sigma rule of thumb that data points with an absolute difference from more than $3\sigma$ are extremely rare and define the boundary of an ellipse using $3\boldsymbol\Sigma'$. Considering the opacity can scale down the size of the ellipse, reducing the area of the AABB and the number of key-value pairs.
When opacity is sufficiently low, the overestimation problem is more significant. As shown in Figure \ref{fig:Opacity_distribution}, most Gaussian opacities in the smallcity scene of the MatrixCity dataset are quite low ( 65.1\% Gaussians have the center opacity less than 0.35), further proving the importance of considering opacity when defining the ellipse.

\paragraph{The defined AABB of the ellipse covers more tiles than the real one.} The preprocessCUDA operation defines the assigned AABB as the bounding box of the circle whose radius is the semi-major axis of the projected ellipse. The AABB is larger than the real AABB of the ellipse, especially for the ellipsoids with high flattening, which introduces computational redundancies for key-values binning, sorting, and rendering.

\paragraph{The duplicateWithKeys kernel assigns redundant tiles in the AABB to key-value pairs} All tiles in AABB are traversed and binned to key-value pairs, even if the tile utterly not covered by the ellipse, which will result in redundant computation in subsequent sorting and rendering process.

\subsection{Observation 2: Inappropriate execution pipelines reduce hardware utilization.}
We find that the main steps in 3DGS exhibit an imbalance between computation and memory access, which can put excessive pressure on the corresponding hardware units, causing computation or memory bottlenecks.
The preprocess step (preprocessCUDA in Algorithm~\ref{preprocessCUDA_algorithm}) kernel has a linear complexity $O(P)$, where P is the number of Gaussians. And each Gaussian is related to a large array of spherical harmonics coefficients. 
This results in significant pressure on cache bandwidth due to the massive read-write operations (lines 17-22 of Algorithm~\ref{preprocessCUDA_algorithm}).

\begin{algorithm}[h]
  \caption{preprocessCUDA
  \label{preprocessCUDA_algorithm}} 

  \begin{algorithmic}[1]
\Require
      $P$,
      $\mathbf{cam}_{pose}$, $\mathbf{s}$,
      $\mathbf{R}$, $\sigma$, $shs$,
        $viewmatrix$, $projmatrix$,
      $\mathbf{o}$
\Ensure
      $radii$, $\mathbf{x}'_{I},depths,\boldsymbol\Sigma$, $\mathbf{c}$,
      $conic\_opacity$, $tiles\_touched$
 \If{$idx \geq P$}
      \State \Return
    \EndIf

    \State $p\_view \gets {in\_frustum}()$
    \If{$!p\_view$}
      \State \Return
    \EndIf

    \State \textcolor{olive}{$p\_proj \gets {{project}}()$}
    \State $\boldsymbol\Sigma \gets {computeCov3D}()$
    \State \textcolor{olive}{$\boldsymbol\Sigma' \gets {computeCov2D}()$}

    \State $\lambda_1, \lambda_2 \leftarrow {eigenvalues}(\boldsymbol\Sigma')$
    \State $radii \gets \lceil 3 \times \sqrt{\max(\lambda_1, \lambda_2)} \rceil$
    \State \textcolor{olive}{$\mathbf{x}'_{I} \gets$ ndc2Pix()}
    \State \textcolor{olive}{$rect\_min,rect\_max \gets {getRect}()$}
    \If{$(rect\_max.x - rect\_min.x) \times (rect\_max.y - rect\_min.y) == 0$}
      \State \Return
    \EndIf

    \State $\textcolor{red}{\mathbf{c}[idx] \gets {computeColorFromSH}()}$
    \State $depths[idx] \gets p\_view.z$
    \State $radii[idx] \gets radii$
    \State $\mathbf{x}'_{I}[idx] \gets point\_image$
    \State $conic\_opacity[idx] \gets (conic, opacities[idx])$
    \State $tiles\_touched[idx] \gets (rect\_max.y - rect\_min.y) \times (rect\_max.x - rect\_min.x)$
  \end{algorithmic}
\end{algorithm}

During the rasterization phase, 3DGS first projects Gaussian spheres on tiles to obtain the tile-gaussian pairs (DuplicateWithKeys). 
Its asymptotic complexity depends on the number of Gaussians and the number of tiles covered by their projection $O(P+N)$. As shown in Algorithm~\ref{duplicatewithkeys_algorithm}, in this step, aside from some bitwise operations, all the operations involve reading and writing to global memory.

\begin{algorithm}[h]
  \caption{duplicateWithKeys
  \label{duplicatewithkeys_algorithm}}

  \begin{algorithmic}[1]
   \Require
      $P$,
      $\mathbf{x}'_{I}$,
      $depths$,
      $offsets$\textcolor{blue}{(Prefix sum of tile counts for each Gaussian)},
      $radii$
    \Ensure
       $gaussian\_keys\_unsorted$\textcolor{blue}{(Array for storing key-value pairs)},
    $gaussian\_values\_unsorted$\textcolor{blue}{(Array for storing key-value pairs)},
\State $idx \gets$ thread\_rank()
    \If{$idx \geq P$}
      \State \Return
    \EndIf
    \If{$radii[idx] > 0$}
      \State $off \gets (idx == 0) ? 0 : offsets[idx - 1]$  
    \State $rect\_min, rect\_max \gets $getRect()
    \For {$y$ of $rect\_min.y$ to $rect\_max.y$}
        \For{$x$ of $rect\_min.x$ to $rect\_max.x$}
          \State \textcolor{purple}{$gaussian\_keys\_unsorted[off] \gets key$ }
          \State \textcolor{purple}{$gaussian\_values\_unsorted[off] \gets idx$} 
          \State $off \gets off + 1$
        \EndFor
      \EndFor
    \EndIf
  \end{algorithmic}
\end{algorithm}

During rendering, the extensive pixel-level calculations introduce computational bottlenecks, whereas the aforementioned two steps are primarily memory access operations. This results in an imbalance between computation and memory access in the entire 3DGS execution workflow, which can lead to under-utilization of GPU hardware resources.


\section{FlashGS Design\label{sec:design}}
\begin{figure*}[t]
    \centering
    \includegraphics[width=\linewidth]{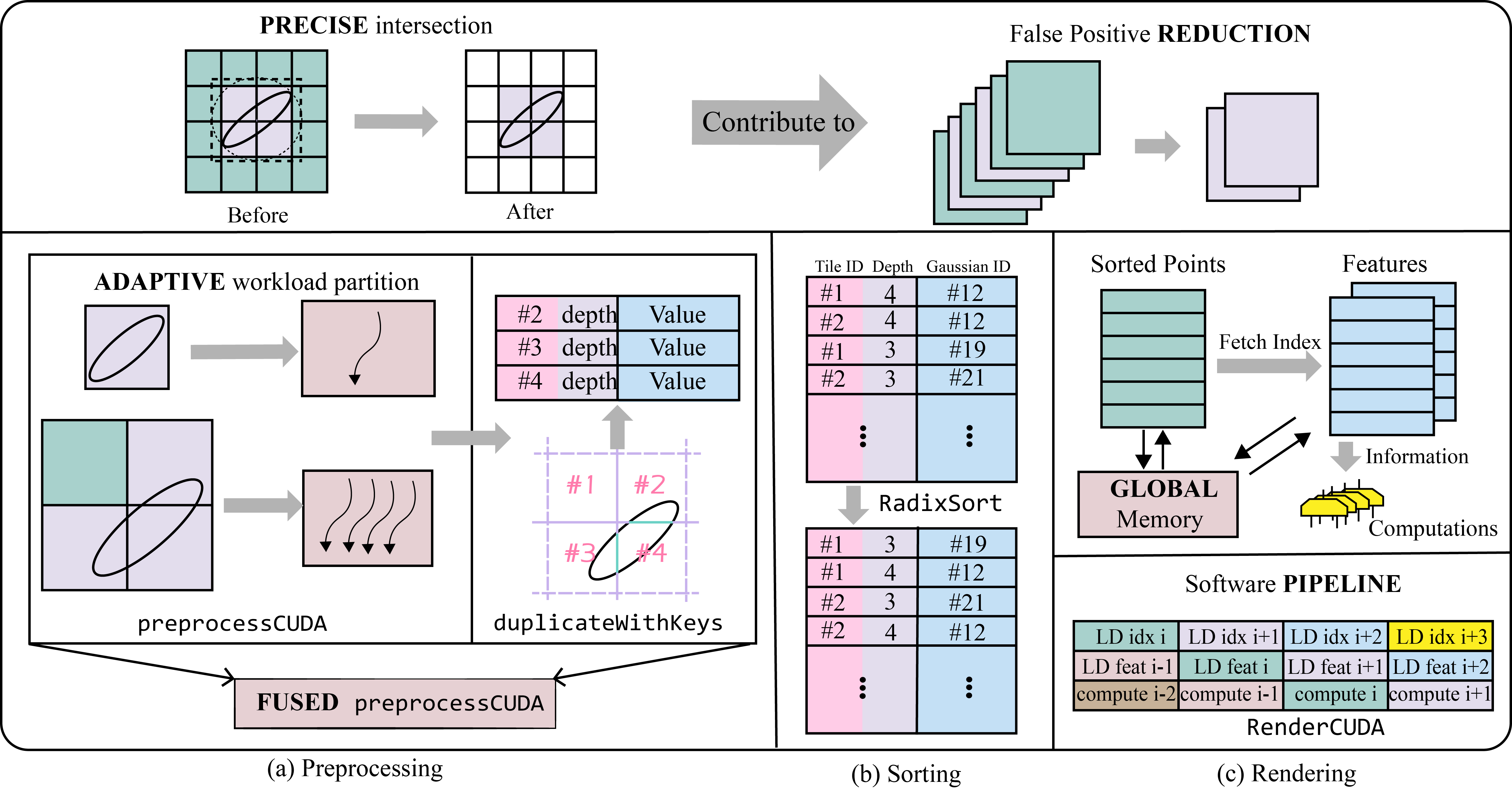}
    \caption{FlashGS Overview.}
    \label{fig:overview}
\end{figure*}
To address the aforementioned issues, we optimize the Gaussian rendering process in two stages. First, we design a more precise intersection algorithm to alleviate subsequent computational bottlenecks related to redundant Gaussian calculations, and use geometric simplifications to avoid expensive intersection computations. Additionally, we reorganized and restructured the computation process based on the characteristics of each kernel/micro kernel to prevent unbalanced workloads. As Figure~\ref{fig:overview} illustrates, the whole process can greatly benefits from the reduced number of false positive tile kv pairs in FlashGS, as well as from the optimization in algorithms and scheduling for runtime.

\subsection{Efficient and Precise Intersection Algorithm\label{sec:design-intersect}}

\begin{figure}[htbp] 
    \centering 
    \includegraphics[width=\linewidth]{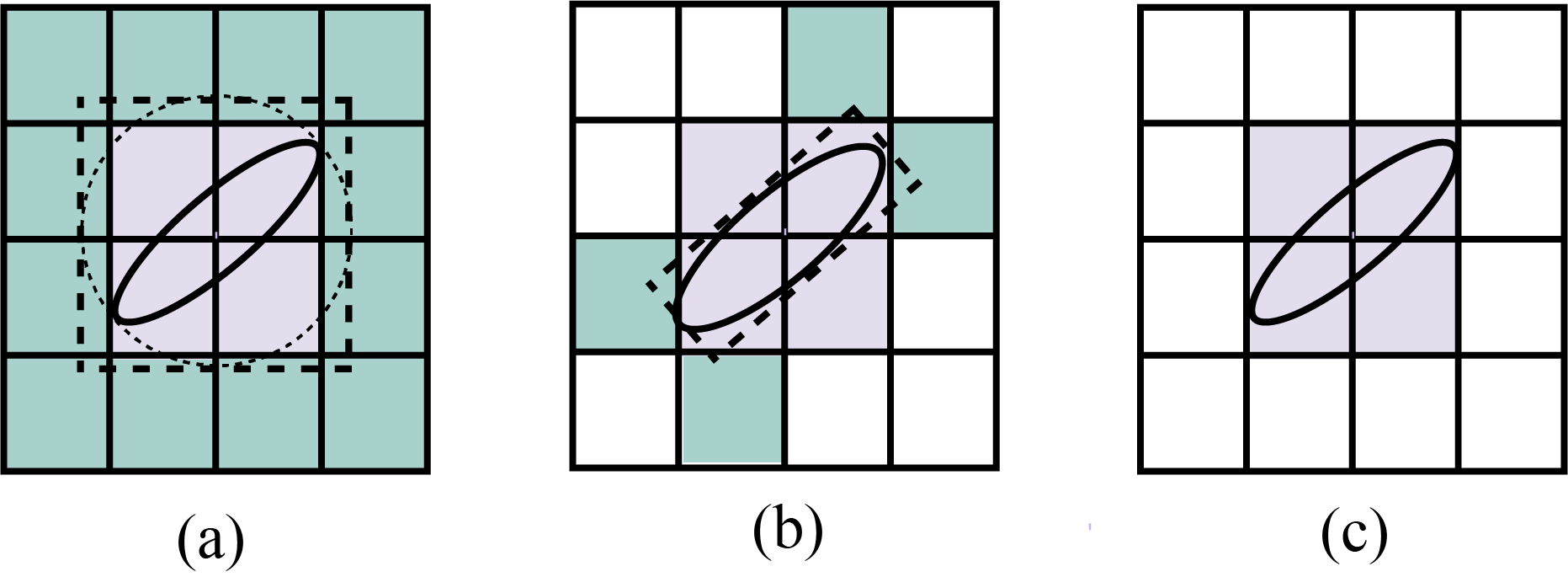} 
    \caption{intersection tiles with the ellipse (purple represents the real intersection tiles, green shows the tiles each method treats as intersected, and white means not). (a) 3DGS uses AABB and gets 16 tiles. (b) GScore applies OBB and gets 8 tiles. (c) Precise intersection shows only 4 tiles.}
    \label{fig:intersection} 
\end{figure}
As shown in Figure~\ref{fig:intersection}, the precise intersection algorithm can reduce the number of Gaussians involved in subsequent computations, thereby alleviating the computational and memory access intensity/pressure. This is crucial because for the two vital bottlenecks, sorting and rendering, their workload and the number of Gaussians to be processed are proportional.

We need to confirm the valid range of the ellipse projected by the Gaussians before performing the precise intersection calculation. As mentioned in Section~\ref{sec:ob1}, for 3D image reconstruction, we must consider not only the effect of the general Gaussian distribution on the effective range but also the changes in opacity. 


3D Gaussian Splatting applies the three-sigma rule to get the valid ellipse:
\begin{equation}\label{equ1}
    r = 3\sqrt{\lambda},
\end{equation}
where $\lambda$ represents the semi-major axis length of the projected ellipse, which corresponds to the eigenvalue of the covariance matrix, and $r$ is the radius of the circle abstracted from the ellipse. 
Considering the statistics in Section~\ref{sec:ob1}, 65\% of Gaussian centers have opacity in quite a low range ([0, 0.35], this range will be explained in the following paragraphs), which motivates us to incorporate opacity $\alpha$ to obtain the effective range:

\begin{equation}\label{equ2}
    \alpha = \alpha_0 \times \text{exp}(-\frac{r^2}{2\lambda}),
\end{equation}
where $\alpha_0$ is the initial (central) opacity of the Gaussian. When $\alpha$ decays to a certain threshold $\tau$, we consider this to be the boundary of the ellipse, as described in Equation~\ref{equ3}:
\begin{equation}
    \label{equ3}
    r = \sqrt{2ln(\frac{\alpha_0}{\tau})\lambda},
\end{equation}

Given that RGB typically has 256 discrete values, typically $\tau = \frac{1}{255}$. 
We can solve that when $\alpha_0 \leq 0.35$, our method outperforms the three-sigma rule, and still use 3$\sigma$ conversely. 

We adopted an exact intersection algorithm to eliminate invalid Gaussians. Consider that a rough intersection algorithm can be highly inefficient for ellipses since they can be quite elongated, which potentially introduces a large amount of unnecessary computations on invalid Gaussians for tiles.
Directly computing the intersection between the Gaussian-projected ellipse and the tiles on the screen is expensive. We reduce the overhead of direct intersection computation through a two-stage filtering process: (1) Use a bounding box (note:  we calculate a rectangle tightly tangent to the ellipse, rather than a square tangent to the circle enclosing the ellipse as in the original algorithm) is to set a coarse-grained range. (2) For tiles within or intersecting the bounding box, check whether they intersect with the ellipse. 

\begin{figure}[htbp] 
    \centering 
    \includegraphics[width=0.7\linewidth]{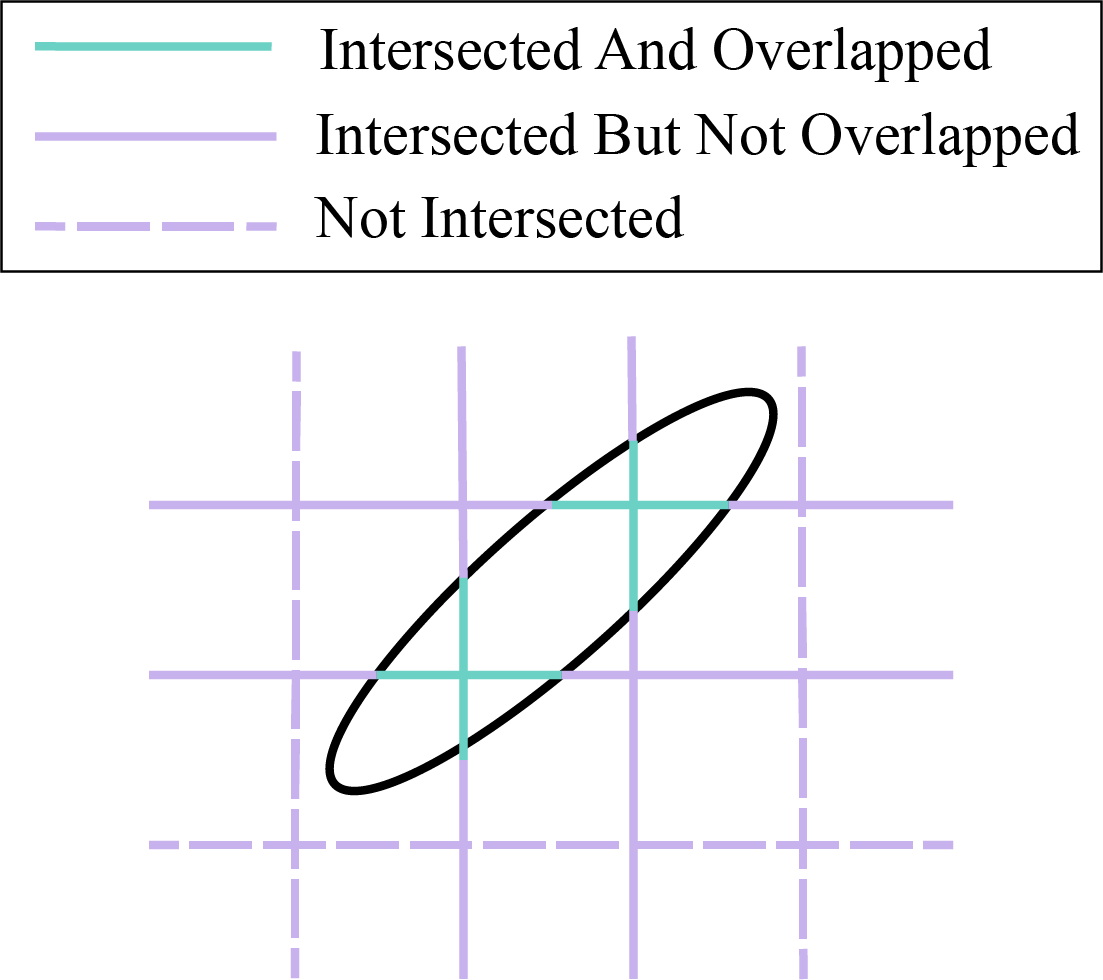} 
    \caption{Geometric simplification for precise per-tile intersection. A tile is considered intersected if the segment of the ellipse intersecting the line of the tile's edge coincides with the edge.} 
    \label{fig:geo-intersection} 
\end{figure}

For each tile intersection, we use geometric equivalent transformations to avoid the cost of directly solving multiple quadratic equations, considering that rectangular tiles are not easily represented by simple equations. 
Our key observation is that the intersection between an ellipse and a rectangle can be transformed into a problem of determining whether the projection line segment of the ellipse onto the line of each edge of the rectangular overlaps with the line segment of the rectangle's edge, as illustrated in Figure~\ref{fig:geo-intersection}. If they intersect, we find the endpoints of the segment on the line that is enclosed by the ellipse. We then use these endpoints to determine whether this segment overlaps with the corresponding edge of the rectangle.

\begin{algorithm}[h]
  \caption{Algorithm to Determine if a Rectangle Tile Needs Rendering}
  \label{algo:intersection} 
  \begin{algorithmic}[1]
    \Require
        Rectangle $R$, Ellipse $E$
    \Ensure
        Boolean indicating if the tile needs rendering
            \If{the center point of $E$ lies in $R$}
                \State \Return true
            \Else
                \For {edge $e$ of $R$}
                    \State $l \leftarrow \text{line containing segment } e$ 
                    \If{$l \cap E \neq \varnothing$ and intersection(s) lie on $e$}
                        \State \Return true
                    \EndIf
                \EndFor
            \EndIf
            \State \Return false
  \end{algorithmic}
\end{algorithm}

As shown in Algorithm~\ref{algo:intersection}, we first classify the intersection between an ellipse and a rectangle into two cases. 
The first case is straightforward: if the center of the ellipse is inside the rectangular, there must be an overlap. 
For the more complex case where the ellipse's center is outside the rectangle, we solve for the intersection of the ellipse with each line containing the edge of the rectangle.


\subsection{Refined and Balanced Workflow\label{sec:design-workflow}}
The workflow of 3DGS rendering is a complex process involving multiple steps as mentioned above. Moreover, different sub-components may have varying computational patterns, and these sub-computations are not entirely independent of each other. Therefore, we need to perform a systematic analysis and optimization from a more holistic perspective.

Our key idea is to balance the computational cost of each part as much as possible to avoid potential bottlenecks. 
Our approach behind this idea is to amortize different types of operations over the timeline to prevent a specific transaction from bursting and putting excessive pressure.
We extract the computational bottlenecks and memory access bottlenecks from different operators and interweave them to form a more efficient execution workflow.

\begin{figure}[htbp] 
    \centering 
    \includegraphics[width=1\linewidth]{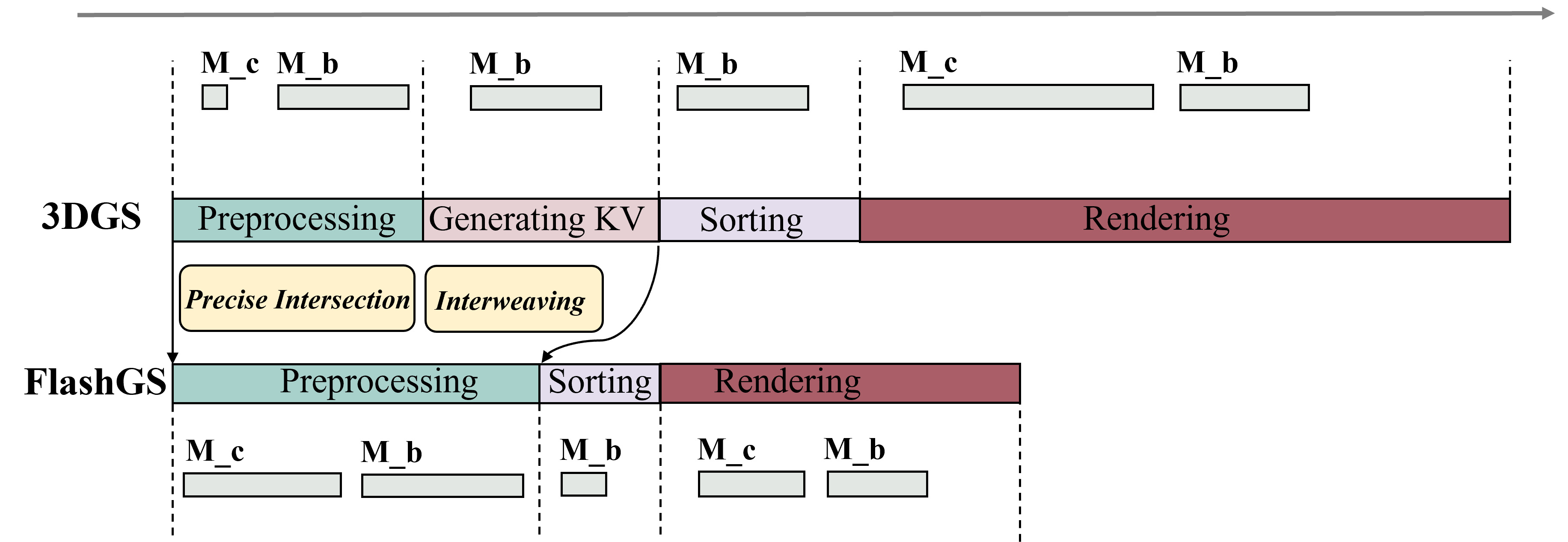} 
    \caption{Schematic for the workflow of original 3DGS and the improved FlashGS. We balance the computation and memory access across various stages and reduce redundant operations (M\_c and M\_b are indicative of the amount of computation and memory access, respectively).} 
    \label{fig:worflow} 
\end{figure}

We observe that before performing rendering, there are several important operations, including the previously mentioned precise intersection, calculating and recording some Gaussian-related information, and constructing key-value pairs for sorting. 
In baseline, both these two functions are bandwidth-limited and involve extensive memory read and write operations. 
However, we observed that the subsequent rendering process is computation-bound. This is due to the unrefined method for Gaussian intersection in the baseline algorithm, which led to massive false positives. Therefore, we may be able to afford more computation during the preprocessing stage. This not only helps to balance the bandwidth-limited parts but also reduces the overhead for subsequent per-pixel rendering. This is one of the key motivations for adopting a precise intersection algorithm mentioned before. 

In the preprocessing stage, we checked the visibility of the Gaussian and then computed and stored various pieces of information needed for subsequent steps, such as depth, Gaussian radius, and coordinates. In the original algorithm, each thread processing a Gaussian that intersects with a tile accesses global memory to read the sphere coefficients and write several Gaussian information at the end of the preprocessing. This results in significant bandwidth contention, causing the workflow to be stalled during global memory access due to implicit synchronization between the separately launched kernels. We combined our new exact intersection algorithm with DuplicateWithKeys() to precisely generate the valid Gaussian key-value pairs.

As previously mentioned, accurately computing the intersection between a tile and a projected ellipse is complex and requires solving multiple equations. Consequently, many operations during this process are compute-bound, leaving surplus bandwidth resources available. Therefore, we reorganized these two operators. By interleaving our exact intersection computations within the delays caused by extensive global memory writes, we ensure that both the GPU's compute units and memory bandwidth are efficiently utilized throughout the execution workflow. Our approach achieves a high compute-to-memory ratio. As the result shown in Figure~\ref{fig:worflow}, the precise intersection operation reduces redundant Gaussians, alleviating the significant memory access overhead during the sorting stage and the additional computations during the rendering stage. In our new preprocessing stage, we can reuse some intermediate results by kernel fusion, transitioning directly from intersection to key-value pair generation, thereby reducing the number of memory read and write operations. Although the total number of computations and memory access operations of the precise intersection and the merged operators may increase, we expose more scheduling opportunities to the compiler and hardware to achieve better balance and amortize over a longer timeline.
\section{FlashGS Implementation}


In this section, we efficiently implement the proposed FlashGS algorithm in Section~\ref{sec:design} to fully leverage algorithm benefits while avoiding the additional overhead of complex computations. 
In our implementation, we further enhance the hardware utilization at runtime by applying 
optimizations on computation, scheduling, and memory management.

\subsection{Preprocess}
In the new preprocessing stage as shown in Algorithm~\ref{algo:opt_preprocessCUDA}, 
Our precise per-tile intersection algorithm with the ellipse introduces massive complex calculations, which might significantly hinder applying this algorithm in current SOTA works. 
It is essential to implement this algorithm efficiently, despite our efforts in Section~\ref{sec:design-workflow} to balance the compute-to-memory ratio by binding the precise intersection with some memory access operations. 
We optimized the algorithm in two key aspects: 1) applying algebraic simplifications to reduce the overhead of a single tile-ellipse intersection; 2) proposing an adaptive scheduling strategy to balance the intersection tasks of the entire projected ellipse of a Gaussian.

\begin{algorithm}[h]
  \caption{preprocessCUDA}
  \label{algo:opt_preprocessCUDA}
  \begin{algorithmic}[1]
    \Require
      $\mathbf{o}, \sigma, shs,viewmatrix, projmatrix,\mathbf{o}_{cam}$,
      $W,H,tan\_fovx,tan\_fovy,focal\_x,focal\_y$,
    \Ensure
      $\mathbf{x}'_{I}$,$offset$,$depths, \boldsymbol\Sigma,\mathbf{c},conic\_opacity$,
      $gaussian\_keys\_unsorted$,
      $gaussian\_values\_unsorted$

    \If{$idx\_vec < P$}
      \State Initialize $p\_orig,$p\_view
      \If{$p\_view.z > 0.2$ \textbf{and} $opacity > 1.0 / 255$}
        \State \textcolor{olive}{$p\_proj \gets$ project()}
        \State \textcolor{olive}{$\boldsymbol\Sigma' \gets$ computeCov2D()}
        \State \textcolor{olive}{$\mathbf{x}'_{I} \gets$ ndc2Pix()}
        \State \textcolor{olive}{$rect\_min, rect\_max \gets$ getRect()}
        \State $point\_valid \gets$ (bounding box is non-zero)
      \EndIf
    \EndIf

    \If{$point\_valid$ \textbf{and} \textcolor{blue}{bounding box is single tile}}
      \If{tile intersects ellipse or contains center}
        \State $key \gets$ compute tile key
        \State $offset \gets$ atomicAdd($curr\_offset$, 1)
        \State \textcolor{purple}{$gaussian\_keys\_unsorted[offset] \gets key$}
        \State \textcolor{purple}{$gaussian\_values\_unsorted[offset] \gets idx\_vec$}
      \EndIf
    \EndIf
    \For{each thread in warp}
      \If{$\textcolor{blue}{multi\_tiles}$}
        \State $parameters \gets$ shuffle values
        \For{$y \gets rect\_min.y$ to $rect\_max.y$}
          \For{$x \gets rect\_min.x$ to $rect\_max.x$}
            \If{$valid$}
                \State \textcolor{purple}{$key \gets$ compute tile key}
                \State \textcolor{purple}{store key and value}
            \EndIf
          \EndFor
        \EndFor
      \EndIf
    \EndFor

    \If{any thread has valid point}
      \State $\mathbf{c}[idx\_vec] \gets$ computeColorFromSH()
      \State store depth, point\_xy, conic\_opacity
    \EndIf
  \end{algorithmic}
\end{algorithm}

\subsubsection{Algebraic Simplification}

We use algebraic equivalence transformations to avoid high-cost operations (instructions).
In our geometric transformations of intersection in Section~\ref{sec:design-intersect}, to determine if two line segments on the same line overlap, we can compare their endpoint coordinates. Assume the two line segments are [a$_i$, b$_i$] and [a$_j$, b$_j$], they overlap if their intersection is not empty as shown in Equation~\ref{eq:seg_intersect-1}:
\begin{equation}
    min(b_i, b_j) - max(a_i,b_i) \geq 0
    \label{eq:seg_intersect-1}
\end{equation}
Specifically, this is equivalent to checking Equation~\ref{eq:seg_intersect-2} that if the right endpoint (end) of one line segment is after the left endpoint (start) of the other line segment, and if the left endpoint (start) of the line segment is before the right endpoint (end) of the other line segment:
\begin{equation}
    a_i \leq b_j\  \&\& \  a_j\geq b_i 
    \label{eq:seg_intersect-2}
\end{equation}

The process of finding the endpoints of intersecting line segments requires solving quadratic equations, which involves high-cost operations such as division and square roots. Assume an edge of a tile is represented by the line segment [a$_i$, b$_i$], The endpoints of the intersecting line [a$_j$, b$_j$] (if exists) are two roots of the quadratic equation formed by the line (on which the tile edge lies) and the ellipse. As shown in the Equation~\ref{eq:seg_intersect-root}, where $DELTA=B^2 - 4AC$, and $A, B, C$ are the coefficients of the quadratic equation $AX^2 + BX + C = 0$:
\begin{equation}
    a_i \leq \frac{-B+\sqrt{DELTA}}{2A}\  \&\&\  b_i \geq \frac{-B-\sqrt{DELTA}}{2A}
    \label{eq:seg_intersect-root}
\end{equation}
Without loss of generality, let $A > 0$, then this condition can be transformed into Equation~\ref{eq:seg_intersect-final} to avoid the high-cost operations, division, and square roots, where the two expressions $e_1=2Aa_i+B$ and $e_2=2Ab_i+B$:
\begin{equation}
    (e_1 \leq 0\ || \ e_1^2 \leq DELTA)\ \&\&\ (e_2 \geq 0\ || \ e_2^2 \leq DELTA)
    \label{eq:seg_intersect-final}
\end{equation}


\subsubsection{Adaptive Size-aware Scheduling}
\begin{figure}[htbp]
    \centering
    \includegraphics[width=0.75\linewidth]{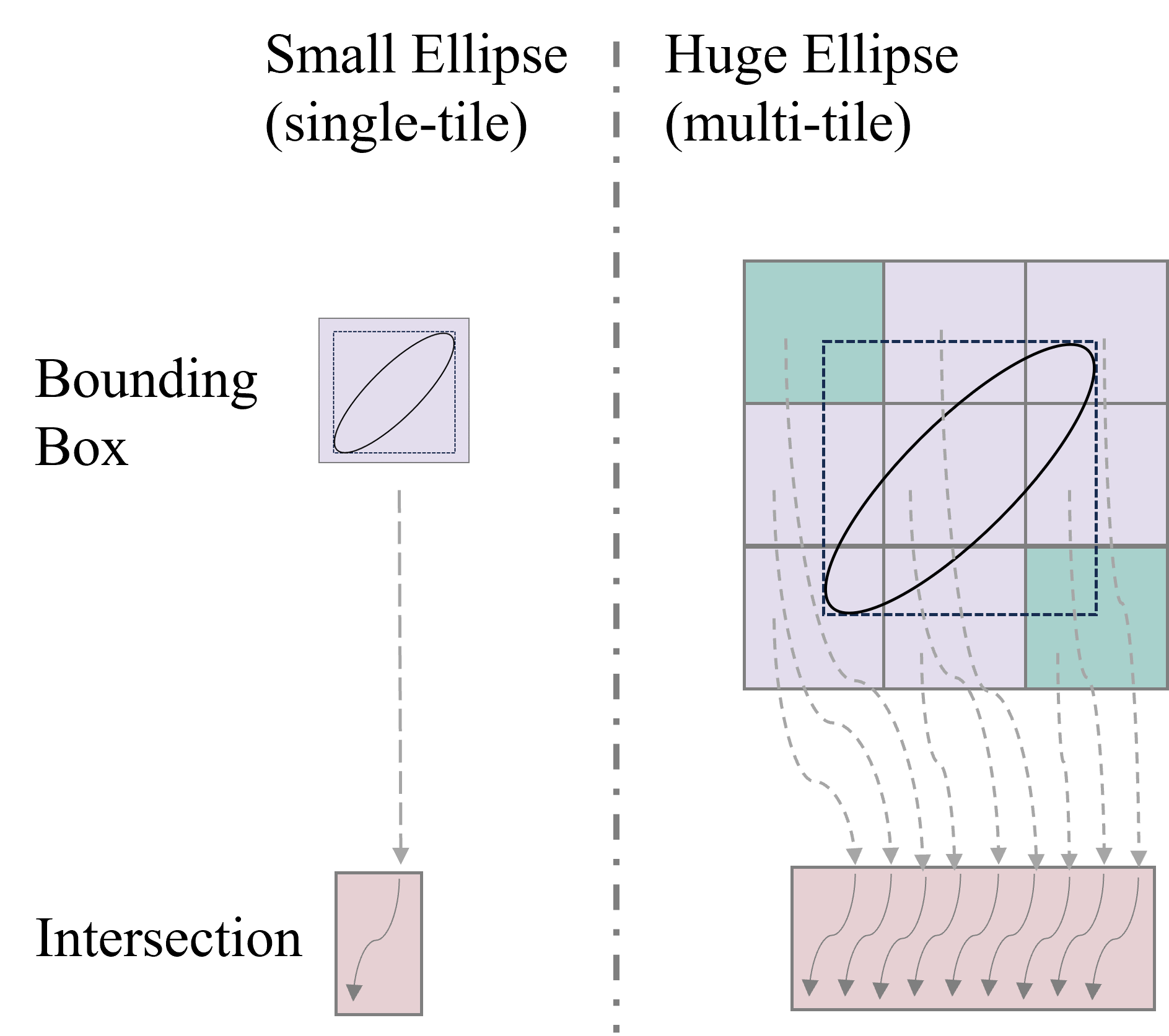}
    \caption{Adaptive task partitioning for Gaussian intersections with varying sizes. If a large ellipse requires processing multiple tiles, other threads within the warp are utilized to collaborate on the intersection.}
    \label{fig:intersect-schedule}
\end{figure}

The precise per-tile intersection algorithm also implies that the workload for intersection tasks can vary significantly for different Gaussians, as the number of tiles covered by ellipses of different sizes can vary greatly. 
In the original 3DGS intersection algorithm, which is based on bounding boxes, this issue does not arise because the covered tile region can be located directly using only the vertices of the bounding box, ignoring the bounding box size.
As shown in the figure~\ref{fig:intersect-schedule}, for Gaussians of different sizes, the resulting bounding boxes, i.e., the number of tiles to be intersected, can vary greatly. 

We automatically select our processing method based on the size of the Gaussians to achieve efficient scheduling. 
To ensure that the workload of each thread is balanced, we use an adaptive mapping. When the ellipse is quite small and only requires calculation for one tile, we continue using the current thread for the precise intersection of this Gaussian. 
When the ellipse is large, we remap the workload to a thread group (in our implementation, we chose a warp to balance synchronization overhead, considering that in practice, the size of the ellipse seldom covers more than 32 tiles). Thus, we achieve an efficient calculation for the entire Gaussian projected ellipse which can intersect with different numbers of tiles. 
After applying all the aforementioned optimizations, our new processing kernel is shown in the Algorithm~\ref{algo:opt_preprocessCUDA}.

\subsection{Rendering}

To address the issue of low hardware utilization during the 3DGS execution, we implement a two-level pipeline optimization from top to bottom. We adjusted the application's execution workflow and the instruction dispatch.

\subsubsection{Low-latency Execution Pipelining}
\begin{figure}[htbp] 
    \centering 
    \includegraphics[width=1\linewidth]{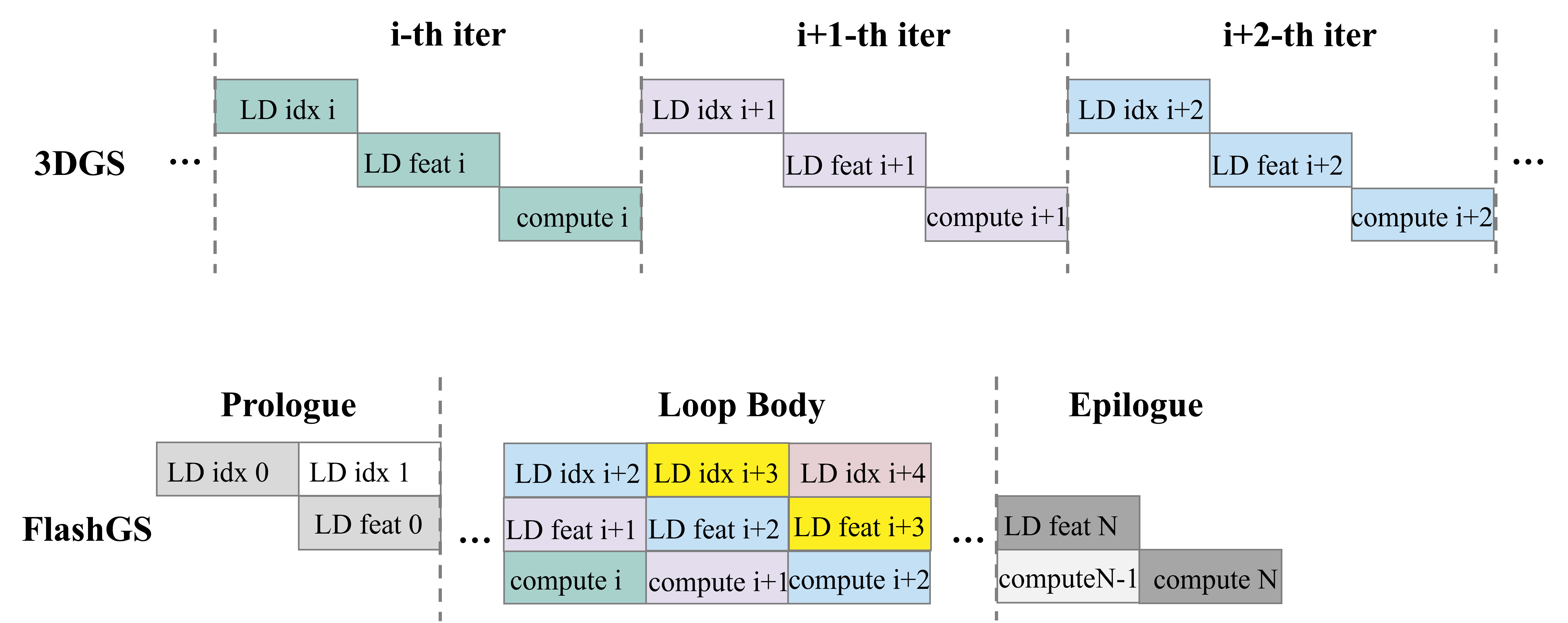} 
    \caption{Software pipelining to achieve better overlap between computation and memory access.
    } 
    \label{fig:pipeline} 
\end{figure}
During rendering, each tile performs computations related to its associated Gaussians, such as calculating transmittance and color. These computations may terminate early based on the accumulated $\alpha$ saturation. These calculations depend on the preprocessed and sorted Gaussian lists, which need to be loaded from global memory in advance. 
The process of reading Gaussian information can be divided into two steps: (1) obtaining the Gaussian index based on the current tile information; (2) using the index to fetch the stored specific information, such as the pixel coordinates of the Gaussian center, the quadratic form of the ellipse, and opacity, among other details.
The high latency of global memory access can hinder the efficient execution of the computation pipeline, especially since there are two dependent steps of global memory access as mentioned above. 

Our core strategy is to issue as many instructions as possible during memory access latency periods to alleviate stalls caused by data dependencies.
Specifically, we introduced a two-level prefetching strategy to maximize the overlap between computation and memory access illustrated in Figure~\ref{fig:pipeline}. Using a software pipelining approach, for the computation at step i, we first fetch the Gaussian index for step i+2 from global memory. Then, using the already fetched index, we retrieve the specific Gaussian information for step i+1 from global memory. Finally, we perform the rendering computation for step i using the Gaussian information fetched at step i-1. We overlap independent instructions as much as possible in our fine-grained rearrangement, thereby achieving a more efficient instruction dispatch pipeline.

\subsubsection{Warp Divergence Control}
Thread divergence within a warp can cause some threads to stall, affecting execution efficiency. The core idea is to minimize conditional branching or move conditional checks to an outer layer (coarser granularity), thus ensuring that threads within a warp execute the same path as much as possible.
In the render step of the original 3DGS algorithm, each thread must check opacity and transmittance when computing a pixel to perform early stopping and determine whether the pixel is valid before writeback. We move the opacity check to the preprocessing stage for effective Gaussian computations. This preemptive filtering at the coarser Gaussian level eliminates redundancy,  replacing the per-pixel conditional checks during rendering.

\subsubsection{Coarse-grained Workload Partition}
We facilitate common data sharing by adjusting the task dispatch granularity so that a single thread handles multiple groups of tasks. 
This larger granularity is mainly based on two considerations. Firstly, it aims to fully utilize fast memory units (such as registers), thus we cannot set the buffer too small (e.g., only buffering a very small amount of data for a single item). Secondly, it takes into account the issue of shared data. In rendering, there is also a certain amount of common computation. We provide a group of tasks as a unit to the compiler to expose more potential sharing opportunities. 
Specifically, we expose the opportunities for common subexpression elimination (CSE) to further improve computational performance by reducing FLOPs. 

\subsection{Common Optimizations}
In addition to the aforementioned efficient parallel algorithm implementation of some critical kernels, we also applied several general techniques to optimize memory and instruction usage throughout the FlashGS computation, achieving further performance benefits.

\subsubsection{Memory Management}
3DGS involves a large number of memory operations, including read, write, and allocation. 
We need a meticulous design to exploit the GPU's hierarchical memory and the memory bandwidth. 

\textit{We utilize the constant memory to reduce latency.}
We carefully analyze the algorithm and extract the special parameters to efficiently utilize the GPU memory hierarchy. The core idea is to leverage special memory access patterns to maximize the use of data loaded into a faster memory hierarchy.
We use constant memory when available, which is highly beneficial for accessing data shared across all threads. For example, in the preprocessing step, all threads need to read some predefined input information, such as projection matrices and transformation matrices. By leveraging the large parameter passing feature (CUDA 12.1), we pass these data directly as parameters to the kernel rather than pointers to their memory locations. This ensures these parameters are placed in constant memory instead of global memory.
Given that global memory latency is approximately 500 cycles, while constant memory access latency with caching is only around 5 cycles, our data placement strategy not only improves access speed but also reduces bandwidth pressure on the L1 cache caused by frequent global memory accesses.

\textit{We reduce the dynamic memory allocation.}
We avoid the performance reduction of frequent memory allocation, which could be caused by the overhead of system calls and memory fragmentation. We extract the dynamic memory allocation operations and related preprocessing operations to the initial stage. For example, a unified preprocessing can be done in advance for computations involving different viewpoints of the same scenario.

\subsubsection{Assembly Optimization}
We also utilize explicit and implicit optimizations to better leverage efficient instructions supported by the GPUs. 
In Gaussian computations, due to the exponential part of the Gaussian distribution, naive calls can result in high overhead. By converting the direct multiplication in the exponential expression to logarithmic form, we first perform addition and then exponential operations. This approach is beneficial because it allows the compiler to use fused multiply-add (FMA) instructions, which have the same overhead as a single MUL or ADD instruction. 
Additionally, when performing logarithmic and exponential operations, we explicitly specify base 2 because the GPU's Special Function Unit (SFU) hardware instructions are optimized for base 2, allowing for direct and efficient calls of the fast \textit{ex2.approx.ftz.f32} PTX instruction.






\section{Evaluation}

\begin{table*}
\scalebox{0.68}{
\begin{tabular}{c|c|c|c|c|c|c|c|c|c|c|c|c}
\multirow{2}{*}{\diagbox{Metrics}{Dataset}}   &  \multicolumn{2}{c|}{Truck} &  \multicolumn{2}{c|}{Train}  &  \multicolumn{2}{c|}{Playroom} &  \multicolumn{2}{c|}{DrJohnson}  &   \multicolumn{2}{c|}{Matrixcity} &   Rubble         \\    
                           &    \multicolumn{1}{c}{1080p}    &        4k      &    \multicolumn{1}{c}{1080p}     &    4k    &    \multicolumn{1}{c}{1080p}     &    4k    &    \multicolumn{1}{c}{1080p}     &    4k    &    \multicolumn{1}{c}{1080p}     &    4k    &    4608*3456   \\\hline 
AvgTime (FlashGS / 3DGS)           &  2.22 / 8.21 &  3.46 / 24.19 & 1.93 / 7.82  & 3.32 / 12.82 & 1.44 / 6.83 & 2.72 / 10.74  &  1.63 / 9.11 & 2.99 / 28.74 & 3.22 / 20.90 & 4.90 / 66.55 & 6.19
 / 44.11\\
MaxTime (FlashGS / 3DGS)           &  4.44 / 10.18 &  4.65 / 29.87 & 3.48 / 12.82  & 4.22 / 42.73 & 2.92 / 10.74 & 4.69 / 32.11  &  4.95 / 16.92 & 6.91 / 57.67 & 4.77 / 40.74 & 7.25 / 140.52 & 9.32 / 67.57
\\
AvgSpeedup                 &  3.76  &  7.01  &  4.20  &  7.48  &  4.92  &  7.86  &  6.18  &  9.99  &  6.56  &  13.64  &  7.41  \\
MaxSpeedup                 &  4.74  &  8.60  &  6.89  &  11.51 &  7.91  &  11.16 &  14.99  &  21.66 &  13.49 &  30.53  &  14.10 \\
MinSpeedup                 &  2.29  &  5.53  &  2.19  &  4.46 &  3.16  &  5.69 &  3.07  &  6.39 &  3.92 &  7.30  &  5.23 \\
AvgTime (FlashGS on A100)                 &  3.37  &  4.87  &  3.05  &  4.58 &  2.36  &  3.98 &  2.59  &  4.35 &  4.18 &  6.55  &  8.08 \\
\end{tabular}}
\caption{FlashGS average/slowest frame rendering time (ms) and corresponding FPS with speedup relative to 3DGS across different datasets and resolutions on 3090 GPU.
}\label{tab:time_comparison}
\end{table*}

\subsection{Experimental Setup}
\subsubsection{Testbed}

We conduct our experiments on two NVIDIA GPU platforms. Most tests are performed on a consumer-grade RTX 3090 (Ampere architecture, 24GB GDDR6X memory, CUDA Compute Capability 8.6). We use an A100 (Ampere architecture, 80GB HBM2e, CUDA Compute Capability 8.0) data-center GPU to study performance sensitivity. 
We compile our codes with GCC 10.5.0 and NVCC in CUDA 12.0. We have also integrated FlashGS as a callable module in Python to enhance usability.

\subsubsection{Dataset}

We comprehensively test FlashGS and 3DGS on 6 representative datasets, which cover very different resolutions on a total of 11 scenes with hundreds or thousands of frames per scene. The Truck and Train are outdoor scenes from the Tanks \& Temple~\cite{knapitsch2017tanks}. The Playroom and DrJohnson are two indoor scenes in DeepBlending~\cite{hedman2018deep}. 3DGS claims that they achieve real-time rendering with 1080p resolution on these datasets. We use another 2 large-scale and high-resolution datasets which beyond 3DGS's capabilities to render in real-time. 
MatrixCity~\cite{li2023matrixcity} is a comprehensive and large-scale high-quality dataset, comprising thousands of aerial images at a resolution of 1080p. Rubble in Mill19~\cite{turki2022mega} contains high-resolution images captured by drones. We train 30K iterations on each dataset with 3DGS to obtain the Gaussian model for rendering.

\subsubsection{Baseline}

We compare FlashGS with 3DGS~\cite{3DGS} and the recent optimizations on GScore~\cite{lee2024gscore}. Since GScore is implemented as a hardware accelerator unit, we cannot directly test it. They claim that GScore achieves 1.86$\times$ speedup with their intersection and scheduling techniques~\footnote{Their shape-aware intersection test provides 1.71$\times$ speedup and subtle skipping offers an additional 15\% improvement.}.

\subsection{Overall Performance}


As shown in Table~\ref{tab:time_comparison}, the rasterization pipe of FlashGS outperforms the original 3DGS algorithm across all the performance metrics on every dataset with different scenes and resolutions. 
We can always achieve > 100FPS rendering on RTX 3090, even for high-resolution and large-scale datasets. In the slowest frame in Rubble of all datasets, we achieve 107.3 FPS. This demonstrates that FlashGS can perform real-time rendering even in extremely large and high-resolution cases.
Flash GS achieves up to 30.53$\times$ speedup with an average of 12.18$\times$ on the Matrixcity dataset at 4k resolution.  We achieve 7.2$\times$ average speedup on all 11 scenes, while achieving 8.6$\times$ speedup on the 7 large-scale or high-resolution tests, which is 3.87$\times$ of the GScores results~\cite{lee2024gscore}. FlashGS achieves a 2.29$\times$ to 30.53$\times$ speedup for all the frames in different scenes, comprehensively proving the generality of our design and optimization.

\subsection{Runtime Breakdown Analysis}
\begin{figure}[htbp]
    \centering
    \includegraphics[width=1.0\linewidth]{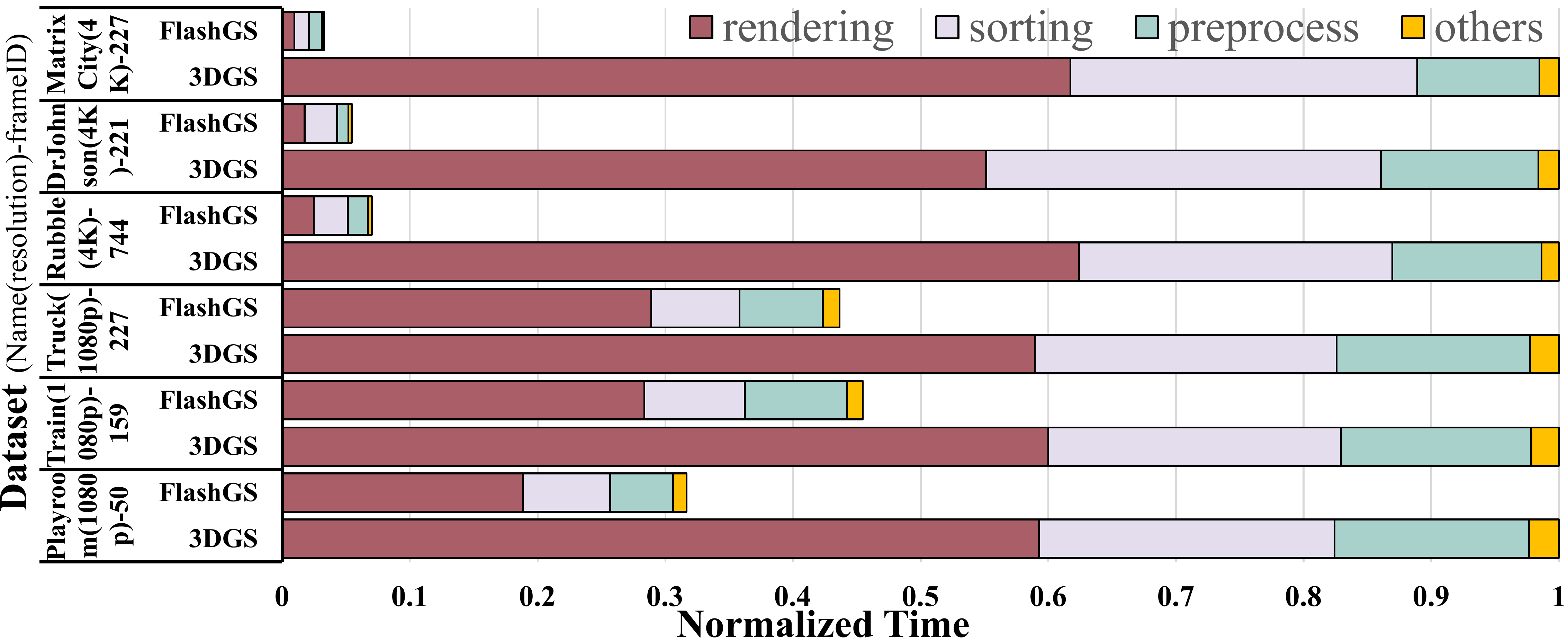}
    \caption{Rasterization runtime breakdown on 6 representative frames from different datasets, normalized to 3DGS.
    }
    \label{fig:breakdown}
\end{figure}
In this section, we compare and analyze the runtime breakdown of FlashGS to 3DGS, aiming to reveal the source of our excellent performance. 
Figure~\ref{fig:breakdown} shows the rasterization time and the breakdown for 6 representative frames shows max or min speedup. This demonstrates that we have accelerated all stages, including preprocess, sort, and render. 
In FlashGS, these stages respectively account for average 19.6\%, 29.1\%, and 47.6\% of the total time, whereas in 3DGS, they account for 13.2\%, 25.4\%, and 59.6\%. The optimizations in the preprocess and render stages have already been discussed earlier, while the speedup in the sorting stage is primarily due to the reduction in the number of key-value pairs to be sorted after applying our precise intersection algorithm. We will give a more detailed analysis below.

\subsubsection{Profiling Results}

\begin{figure}[htbp]
    \centering
    \includegraphics[width=1.0\linewidth]{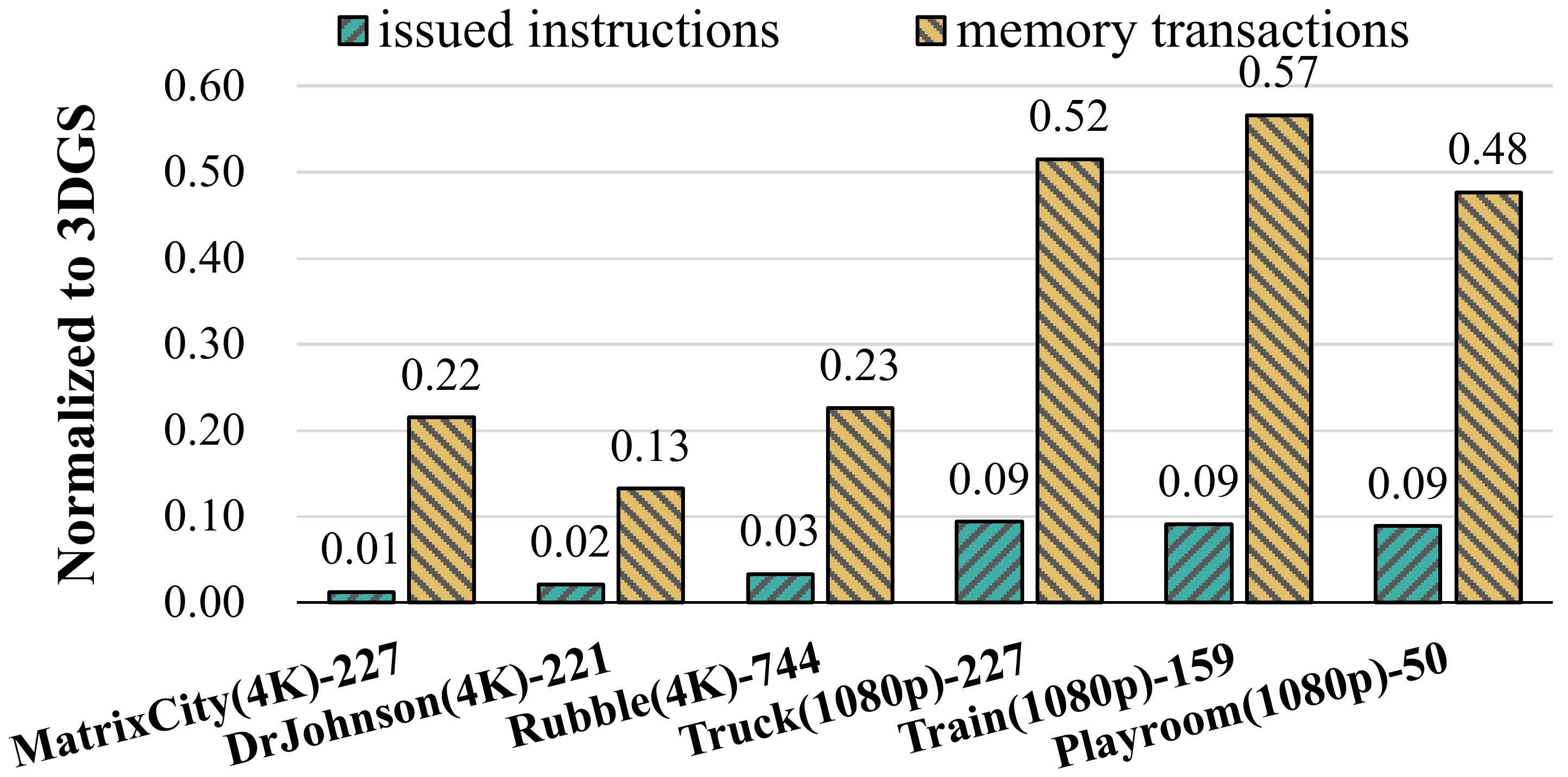}
    \caption{Profiling results of FlashGS: The number of instructions issued in rendering and the memory transactions in preprocessing. All results are normalized to 3DGS.}
    \label{fig:profiling}
\end{figure}

We further demonstrate our optimizations in the rendering and preprocessing stages through profiling results of memory and compute units, as shown in the figures. Figure ~\ref{fig:profiling} shows that we reduce the issued instructions in the rendering stage, alleviating the computation-bound problem. This prove the effectiveness of our precise intersection algorithm and the optimizations for low-latency rendering. The total issues instructions is significantly reduced by one to two orders of magnitude. For memory access trasactions in preprocessing, we also reduced the number of global memory accesses by 43\%-87\% compared to 3DGS.

\begin{figure}[htbp]
    \centering
    \includegraphics[width=1.0\linewidth]{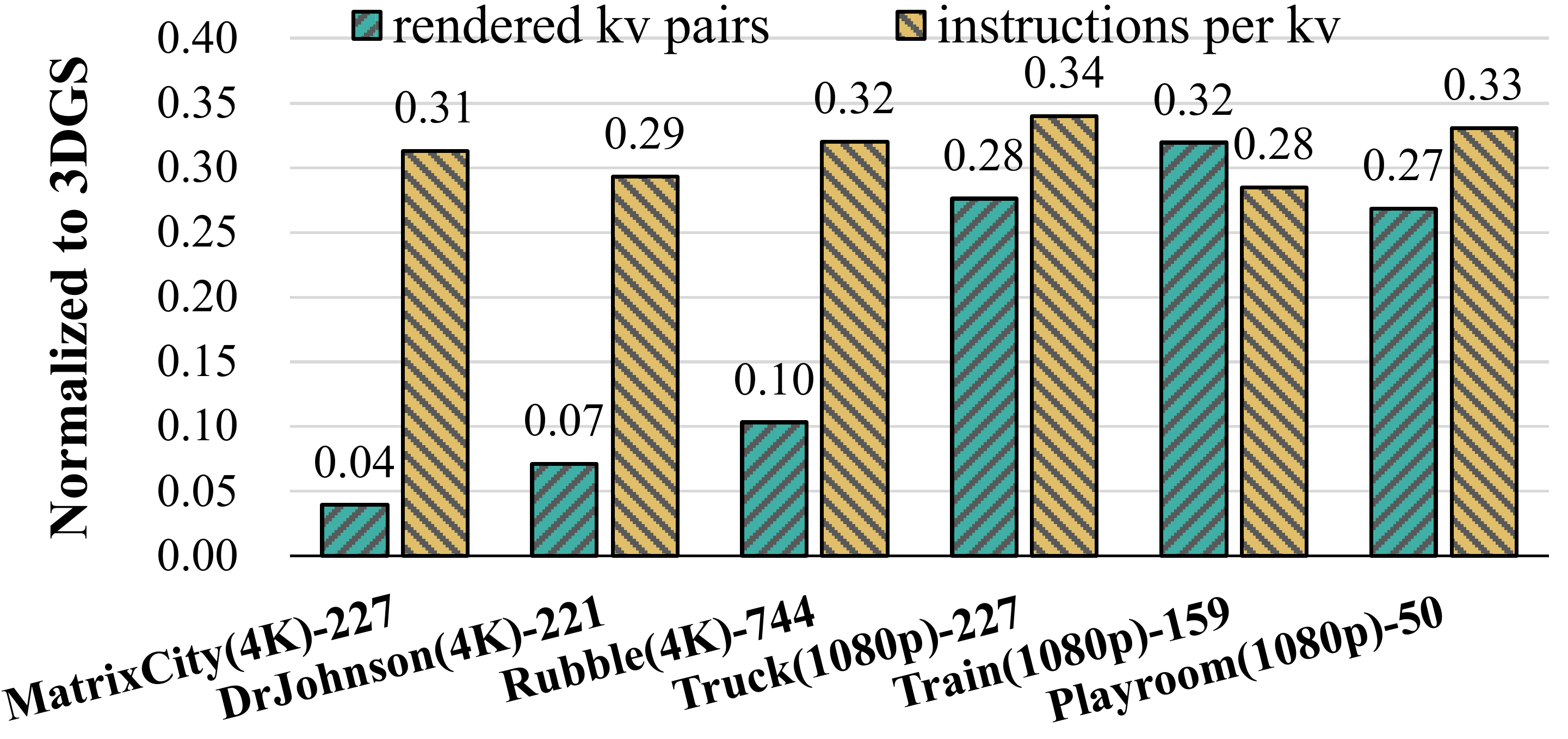}
    \caption{Number of rendered Gaussian-tile (kv) pairs and instructions issued per pair of FlashGS (Normalized to 3DGS).}
    \label{fig:numrender}
\end{figure}

Figure~\ref{fig:numrender} further shows the reason for the significant instruction reduction in the rendering stage, which is dominant in the rasterization. In tile-based rendering, the total number of issued instructions is the product of the number of tiles and the instructions per tile. Therefore, this reduction mainly stems from two aspects. In Figure~\ref{fig:numrender}, the number of rendered key-value pairs benefit from our intersection optimization, reducing by 68\%-96\%. The instructions involved in each tile's computation also decrease by 67\%-71\%, as we further optimize the renderCUDA kernel at the instruction level. The reduction in the number of generated key-value pairs also benefits the sorting process, as it significantly decreases the size of the list to be sorted. Additionally, less memory is required to store these key-value pairs.



\subsection{Sensitivity Study}
In this section, we study the FlashGS performance sensitivity on different GPUs and sciences.
\subsubsection{Performance on A100}
We also conduct the experiments presented in Table~\ref{tab:time_comparison} on the A100 GPU, and the results are shown in Table~\ref{tab:time_comparison}. The total rasterizerization time is slower on the A100 compared to the 3090, with on average 1.43$\times$ slower. This is primarily due to the rendering step is dominant, as shown in Figure~\ref{fig:breakdown}, which heavily relies on FP32 computation, where the FP32 peak performance of the A100 is only 19.5TFLOPS while the 3090 is 35.6TFLOPS. Although the operations before rendering, such as sorting, are mainly memory-bandwidth-bound, and the A100 has a higher bandwidth than 3090, these benefits can not outweigh the render performance reduction. However, we still achieve significant speedups to 3DGS in all datasets on the A100 with an average of 123.8-423.7 FPS across 11 scenes in Table~\ref{tab:time_comparison}, always reaching real-time rendering.

\subsubsection{Performance on different scenes}
In Table~\ref{tab:time_comparison}, we observe two main results: 1) For the same dataset, the speedup at 4K resolution is higher than at 1080p. 2) Across different datasets, the highest speedup is achieved in large-scale, high-resolution scenes (MatrixCity-4K). This is primarily source from our precise intersection calculations, where the computational load in the rendering is proportional to the number of Gaussians and the number of tiles each Gaussian intersects. In large scenes with more objects, the number of Gaussians increases; for high resolutions, the Gaussians become larger to render finer details. 
In such cases, the redundancy elimination effect is more pronounced. And if in a very small and simple scene, each Gaussian might be so small that it intersects only one tile, leading to a lower improvement of our precise intersection.

\subsection{Image Quality}
\begin{figure}[htbp]
    \centering
    \includegraphics[width=1.0\linewidth]{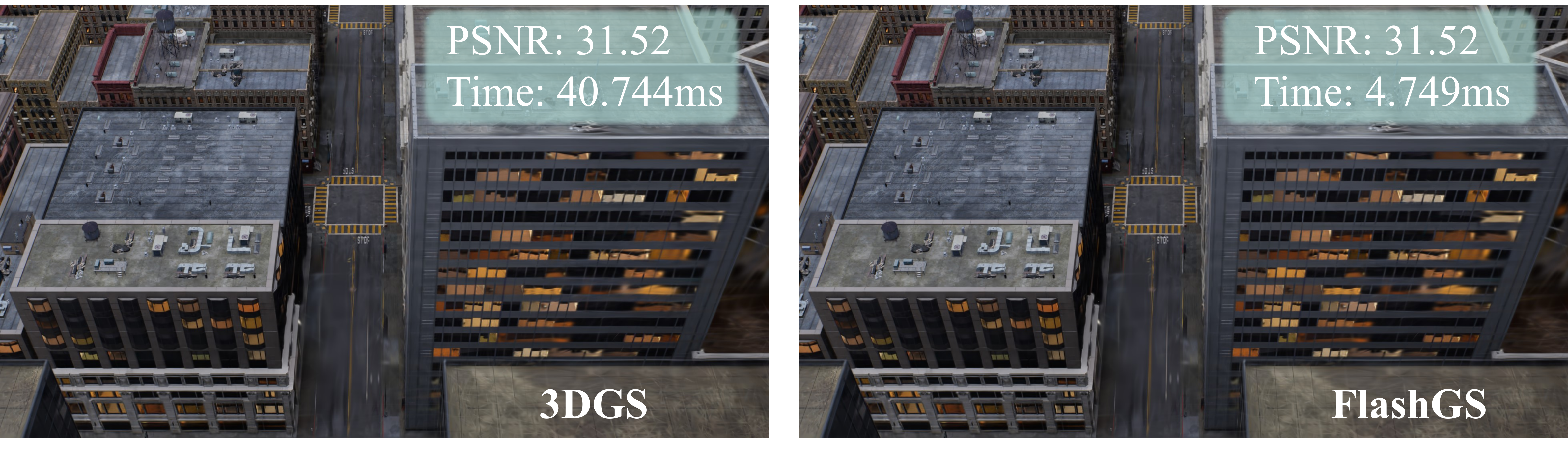}
    \caption{The rendering quality and rasterization time on Matricity(1080p)-800 frame.}
    \label{fig:quality}
\end{figure}
Figure~\ref{fig:quality} shows the most complex frame of the origin large-scale MatrixCity dataset at 1080p, to compare the output quality. We compare the Peak Signal to Noise Ratio (PSNR) between FlashGS and 3DGS, which is a standard metric in computer graphics (higher is better). The result shows that FlashGS does not change the quality, keeping 31.52 PSNR with 8.57$\times$ speedup. 
It is reasonable that our precise intersection algorithms only reduce the false-positive redundancies. And we do not apply pruning or quantization strategies in our implementation so there is no accuracy loss.
\if false
In FlashGS, these stages respectively account for xx\%, xx\%, and xx\% of the total time, whereas in 3DGS, they account for XX\%, XX\%, and XX\%. The optimizations in the preprocessing and rendering stages have already been discussed earlier, while the speedup in the sorting stage is primarily due to the reduction in the number of key-value pairs to be sorted after applying our precise intersection algorithm.

\subsubsection{Profiling Results}
We further demonstrate our optimizations in the rendering and preprocessing stages through profiling results of memory and compute units, as shown in the figures. Figure x shows that we reduce the issue slot busy rate in the rendering stage while increasing the utilization of the FMA (Fused Multiply-Add) units, thereby fully optimizing the compute bottleneck. For the preprocessing stage, Figure x illustrates our mitigation of bandwidth busy, which validates the effectiveness of our inter-kernel redesign discussed in Section~\ref{}, balancing the memory access bottleneck.

\subsection{Sensitivity Study}
In this section, we study the FlashGS performance sensitivity on different GPUs and sciences.
\subsubsection{Performance on A100}
We also conduct the experiments presented in Table~\ref{tab:time_comparison} on the A100 GPU, and the results are shown in Table~\ref{X}. The total rasterizer time is slower on the A100 compared to the 3090. This is primarily due to the render step, which heavily relies on FP32 computation, where the FP32 peak performance of the A100 is only 19.5TFLOPS while the 3090 is 35.6TFLOPS. Although the operations before rendering, such as sorting, are mainly memory-bandwidth-bound, and the A100 has a higher bandwidth than 3090, the render step dominates the rasterizer time. Hence, the A100 is overall slower than the 3090. However, we still achieve significant speedups to 3DGS in all datasets on the A100, reaching real-time rendering.
\fi

In Table~\ref{tab:mem_comparison}, we observe two main results: 1) For the same dataset, the speedup at 4K resolution is higher than at 1080p. 2) Across different datasets, the highest speedup is achieved in large-scale, high-resolution scenes (MatrixCity-4K). This is primarily source from our precise intersection calculations, where the computational load in the rendering is proportional to the number of Gaussians and the number of tiles each Gaussian intersects. In large scenes with more objects, the number of Gaussians increases; for high resolutions, the Gaussians become larger to render finer details. 
In such cases, the redundancy elimination effect is more pronounced. And if in a very small and simple scene, each Gaussian might be so small that it intersects only one tile, leading to a lower improvement of our precise intersection.

\subsection{Image Quality}
\begin{figure}[htbp]
    \centering
    \includegraphics[width=1.0\linewidth]{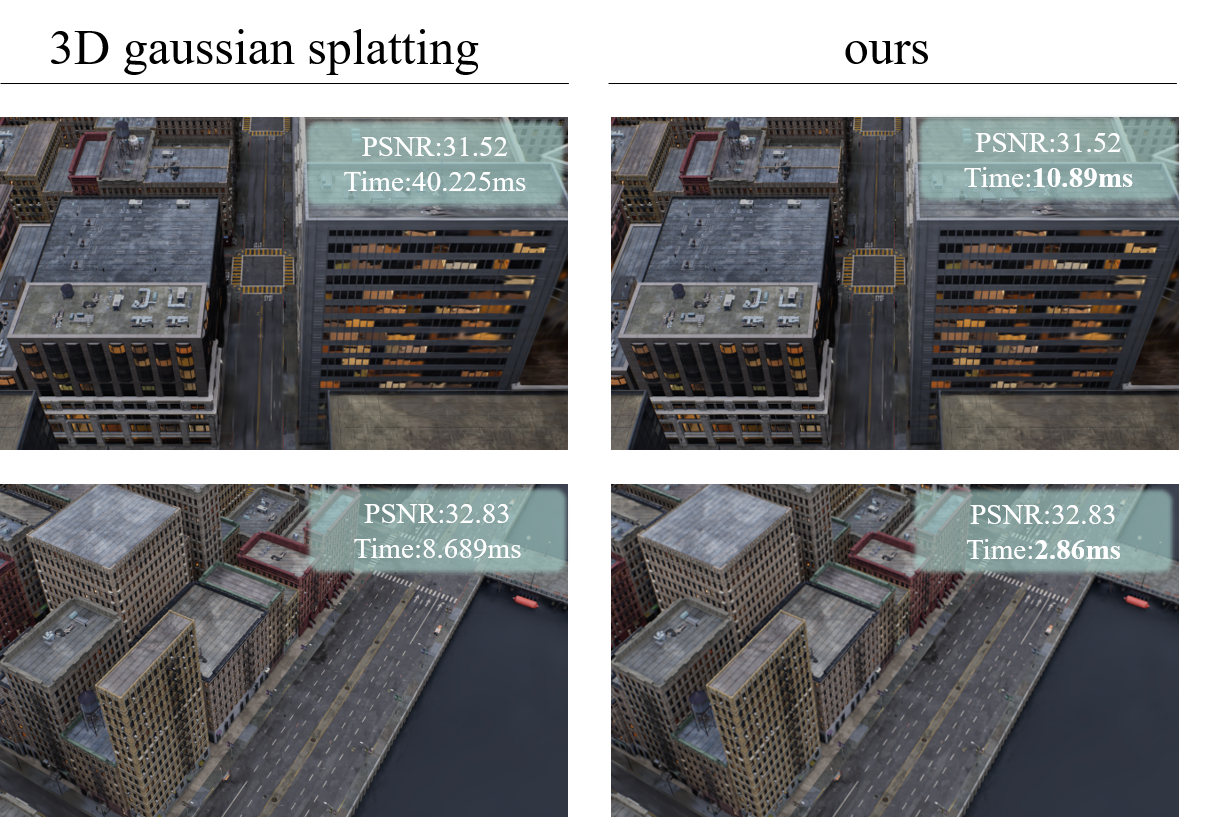}
    \caption{Above is the 800th image (the longest one), below is the 809th image (the shortest one), the left is the baseline, and the right is the  FlashGS (ours)
    }
    \label{fig:quality}
\end{figure}
Figure~\ref{fig:quality} shows two representative frames of the large-scale MatrixCity dataset. We compare the PSNR between FlashGS and 3DGS, and the result shows that FlashGS does not change the quality. It is reasonable that our precise intersection algorithms only reduce the false-positive redundancies. And we do not apply pruning or quantization strategies in our implementation so there is no accuracy loss.

\subsection{Memory Comparison}

\begin{table}
\scalebox{0.74}{
\begin{tabular}{c|c|c|c}
Method                    & Before Rendering &  After Rendering  & \#key-value pairs \\    
\hline 
3DGS                      & 7.86 GB          &  13.45 GB         & 56,148,670 \\
gsplat ($packed$ $False$) & 4.52 GB          &  10.75 GB         & 56,996,302 \\
gsplat ($packed$ $True$)  & 4.52 GB          &  9.83 GB          & 56,998,101 \\
FlashGS                   & 6.83 GB          &  6.83 GB          & 3,436,142 \\                
\end{tabular}}
\caption{Memory allocated before and after rendering the 800th frame in MatrixCity dataset.}\label{tab:mem_comparison}
\end{table}

FlashGS allocates less memory than 3DGS and gsplat, up to 49.2\% reduction. 
Table \ref{tab:mem_comparison} compares the memory usage before and after rendering the 800th frame on an NVIDIA A100 GPU for different models, including gsplat (with $packed$ set to $True$ and $False$), the original 3DGS, and FlashGS. 
Specifically, the gsplat with $packed$ set to $False$ has a maximum memory allocation of 10.75 GB after rendering, while with $packed$ set to $True$, the maximum memory allocation is reduced to 9.83 GB, significantly decreasing memory usage. This reduction occurs because, when $packed$ is set to $True$, the rasterization process is more memory-efficient, packing intermediate tensors into sparse tensor layouts. This is particularly beneficial in large scenes where each camera only sees a small portion of the scene, greatly reducing memory usage. However, this also introduces some runtime overhead.
In contrast, the original GS uses the most memory, with a memory allocation of 13.45 GB after rendering
, making it an important factor to consider when dealing with complex scenes.

FlashGS also ensures consistency and predictability in memory usage through a static allocation method, with a maximum memory allocation of 6.83 GB, which is lower than the other models. This demonstrates its superior efficiency and performance when handling large scenes. The static allocation method effectively avoids fluctuations in memory allocation and release processes, resulting in more stable memory management.

The number of kv pairs generated by each method is a crucial factor in determining memory usage when rendering. The original 3DGS model generates 56,148,670 key-value pairs, which leads to its higher memory consumption. In comparison, gsplat with $packed$ set to $False$ and $True$ generates 56,996,302 and 56,998,101 key-value pairs respectively, showing a slight increase. And it proves that the memory reduction come from its compression techniques. 
FlashGS, on the other hand, generates only 3,436,142 key-value pairs, drastically reducing the memory usage. 




\section{Conclusions}
We propose FlashGS, enabling real-time rendering of large-scale and high-resolution scenes. 
In this paper, we achieved a fast rendering pipeline through a refined algorithm design and several highly optimized implementations, addressing the redundancy and improper compute-to-memory ratio issues present in original 3DGS. 
FlashGS significantly surpasses the rendering performance of existing methods on GPUs, achieves efficient memory management, while maintaining high image quality.
\if false

\begin{acks}                            
  This material is based upon work supported by the
  \grantsponsor{GS100000001}{National Science
    Foundation}{http://dx.doi.org/10.13039/100000001} under Grant
  No.~\grantnum{GS100000001}{nnnnnnn} and Grant
  No.~\grantnum{GS100000001}{mmmmmmm}.  Any opinions, findings, and
  conclusions or recommendations expressed in this material are those
  of the author and do not necessarily reflect the views of the
  National Science Foundation.
\end{acks}

\appendix
\section{Appendix}

Text of appendix \ldots
\fi

\bibliographystyle{ACM-Reference-Format}
\bibliography{./output}

\end{document}